\documentclass[pdflatex,sn-chicago]{sn-jnl}

\usepackage{graphicx}%
\usepackage{multirow}%
\usepackage{amsmath,amssymb,amsfonts}%
\usepackage{amsthm}%
\usepackage{mathrsfs}%
\usepackage[title]{appendix}%
\usepackage{xcolor}%
\usepackage{textcomp}%
\usepackage{manyfoot}%
\usepackage{booktabs}%
\usepackage{algorithm}%
\usepackage{algorithmicx}%
\usepackage{algpseudocode}%
\usepackage{listings}%

\usepackage{wrapfig}
\usepackage{float}
\usepackage{subfigure}
\usepackage{afterpage}
\usepackage{graphicx}
\usepackage{xspace}

\usepackage{booktabs}
\usepackage{xcolor}

\usepackage{blindtext}
\usepackage{cleveref}

\usepackage{anyfontsize}

\makeatletter
\DeclareRobustCommand\onedot{\futurelet\@let@token\@onedot}
\def\@onedot{\ifx\@let@token.\else.\null\fi\xspace}
\def\eg{\emph{e.g}\onedot} 
\def\ie{\emph{i.e}\onedot} 
 
 \def\vs{\emph{vs}\onedot}
 
\def\etal{\emph{et al}\onedot}
\makeatother

\begin{document}

\title[Article Title]{Towards Multi-Modal Animal Pose Estimation: A Survey and In-Depth Analysis}


\author*[1]{\fnm{\textbf{Qianyi}} \sur{\textbf{Deng}}}\email{qianyi.deng@cs.ox.ac.uk}

\author*[1]{\fnm{\textbf{Oishi}} \sur{\textbf{Deb}}}\email{oishi.deb@eng.ox.ac.uk}

\author[2]{\fnm{Amir} \sur{Patel}}\email{amir.patel@ucl.ac.uk}

\author[1]{\fnm{Christian} \sur{Rupprecht}}\email{christian.rupprecht@cs.ox.ac.uk}

\author[1]{\fnm{Philip} \sur{Torr}}\email{philip.torr@eng.ox.ac.uk}

\author[1]{\fnm{Niki} \sur{Trigoni}}\email{niki.trigoni@cs.ox.ac.uk}

\author[1]{\fnm{Andrew} \sur{Markham}}\email{andrew.markham@cs.ox.ac.uk}

\affil[1]{
\orgname{University of Oxford}, 
\country{UK}}

\affil[2]{ 
\orgname{University College London}, \country{UK}}


\abstract{Animal pose estimation (APE) aims to locate the animal body parts using a diverse array of sensor and modality inputs (\eg RGB cameras, LiDAR, infrared, IMU, acoustic and language cues), which is crucial for research across neuroscience, biomechanics, and veterinary medicine. By evaluating 176 papers since 2011, APE methods are categorised by their input sensor and modality types, output forms, learning paradigms, experimental setup, and application domains, presenting detailed analyses of current trends, challenges, and future directions in single- and multi-modality APE systems. The analysis also highlights the transition between human and animal pose estimation, and how innovations in APE can reciprocally enrich human pose estimation and the broader machine learning paradigm. Additionally, 2D and 3D APE datasets and evaluation metrics based on different sensors and modalities are provided.
A regularly updated project page is provided in this \href{https://github.com/ChennyDeng/MM-APE}{GitHub link}.}




\maketitle

\section{Introduction}

Animal pose estimation (APE) identifies the spatial configuration of an animal's body parts using a diverse array of sensor and modality inputs, which is important in numerous disciplines \citep{karashchuk2021anipose}. For instance, neuroscientists rely on accurate measurements of animal movement to correlate with brain dynamics \citep{mathis2018deeplabcut, krakauer2017neuroscience, pereira2020quantifying, luxem2022identifying}. Biomechanists analyse the movement of specific animal body structures to comprehend biomechanical phenomena, such as mechanical properties, stress-strain relationships, and functionality of different body structures \citep{patel2013rapid, joska2021acinoset, patel2014rapid}. In addition, ecologists monitor wildlife behaviours to develop more effective conservation strategies. Similarly, ethologists examine detailed animal gestures and postures to study intra- and inter-species communication and societal structures \citep{zuffi2019three, ng2022animal}. Moreover, veterinary practitioners employ precise APE techniques to diagnose health conditions \citep{farahnakian2021multi, hu20233d} and assess recovery post-treatment, thereby mitigating potential disorders \citep{fang2021pose}.

In this paper, we analyse the current state of deep learning-based APE in both single- and multi-modality approaches. We discuss the developments in single-sensor and single-modality methods, explore the challenges they face, and subsequently study the potential of multi-sensor and multi-modal systems to overcome these limitations and further enhance the accuracy and robustness of APE. To provide context, we first overview the extensive research conducted in Human Pose Estimation (HPE).

\subsection{Human Pose Estimation}

The advancements in HPE provides insights to develop the APE methods due to the skeleton similarity between humans and animals. Therefore, we first provide an overview of HPE to help contextualize the current state and future potential of APE.

The field of HPE has rapidly evolved from manual feature engineering to automated feature learning over the past decade, due to the developments in deep learning approaches. Modern HPE methods are mainly RGB-based and predominantly employ convolutional neural networks (CNNs). In the early stages, models like VGG \citep{simonyan2014very}, ResNet-50 \citep{he2016deep}, ResNet-101 \citep{he2016deep}, and ResNet-152 \citep{he2016deep} were employed as backbones for HPE from images or videos. The stacked hourglass network \citep{newell2016stacked} then presented a novel architecture characterized by repeated bottom-up and top-down processing. This design aimed to capture and consolidate features at multiple resolutions, enabling precise localization of keypoints across scales. Subsequent innovations, such as the High-Resolution Network (HRNet) \citep{wang2020deep} and its variants \citep{cheng2020higherhrnet, yu2021lite}, have further advanced spatial information preservation by maintaining high-resolution representations throughout the entire network. Recently, transformer-based methods \citep{yang2021transpose, yuan2021hrformer} have integrated attention mechanisms to capture long-range dependencies and spatial correlations among joints and appearances, expanding the possibilities of HPE. Beyond vision-based HPE, alternative sensors like inertial measurement units (IMUs) \citep{huang2020deepfuse}, millimeter-wave (mmWave) radar \citep{sengupta2020mm, xue2021mmmesh, 15922} and thermal cameras \citep{smith2023human, chen2020multi} have been explored, tailoring HPE to diverse use cases. Additionally, multi-sensor and multi-modal datasets \citep{ma2024nymeria, yan2023cimi4d, an2022mri} and approaches \citep{furst2021hperl, zheng2022multi, an2022mri} have emerged as an innovative step over unimodal HPE to overcome individual sensor and modality limitations, offering a synergistic fusion of data that promises increased robustness and precision, especially in scenarios marked by occlusions, changing lighting conditions, and dynamic environments.

\subsection{Definition of Animal Pose Estimation}\label{APE_Definition}

\begin{figure*}[!t]
\centering
\subfigure[\textbf{Keypoints Pose Representations.} (a) A 2D keypoints pose representation example adapted from \citep{Nath2019}. (b) A 3D keypoints pose representation example adapted from \citep{joska2021acinoset}.]{\includegraphics[width=8cm, height = 3.9cm]{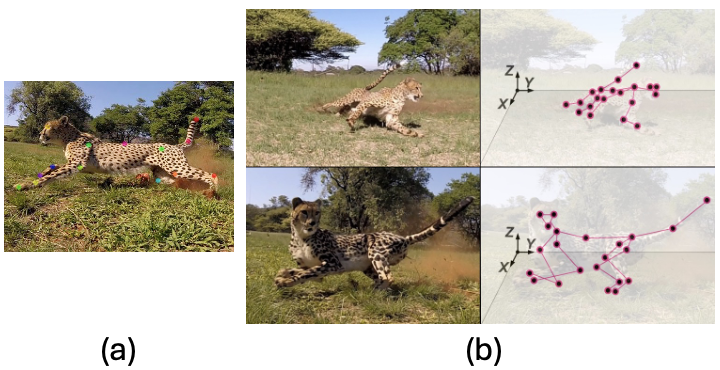}%
\label{fig:Skeleton Pose Representations}}
\hfil
\subfigure[\textbf{Body Mesh Reconstructions.} (a) Original RGB image. (b) Reconstructed mesh for (a) using the 3D model-based body mesh reconstruction adapted from \citep{zuffi20173d}. (c) Target RGB image. (d) Reconstructed mesh for (c) applying the 3D model-free body mesh reconstruction method from MagicPony \citep{wu2023magicpony}.]{\includegraphics[width=8cm, height = 3.9cm]{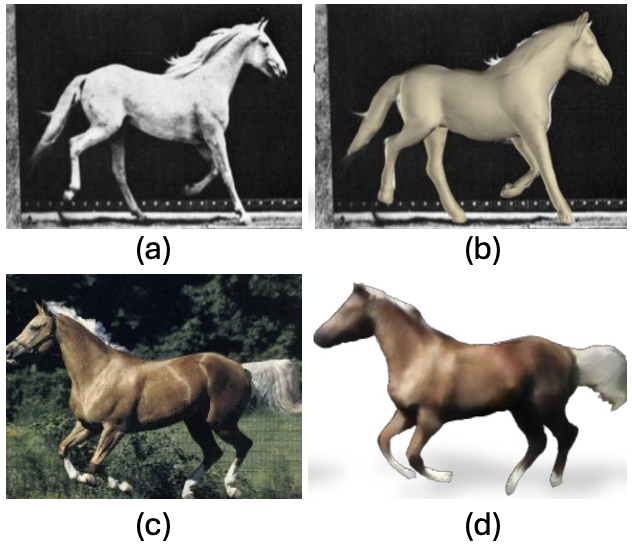}%
\label{fig:Body Mesh Reconstructions}}
\caption{Different animal pose estimation methodologies. \centering} 
\label{fig_APEkpmesh}
\end{figure*}

APE, as discussed in this review, is defined as the task of identifying and localizing body joints of terrestrial mammals using a diverse array of sensor inputs. This task results in the reconstruction of the animal's pose, depicted as either a skeletal representation marked by keypoints (\Cref{fig:Skeleton Pose Representations}) or as a comprehensive body mesh (\Cref{fig:Body Mesh Reconstructions}).

\begin{figure*}[!t]
\centering
\includegraphics[width=\linewidth, height = 3.9cm]{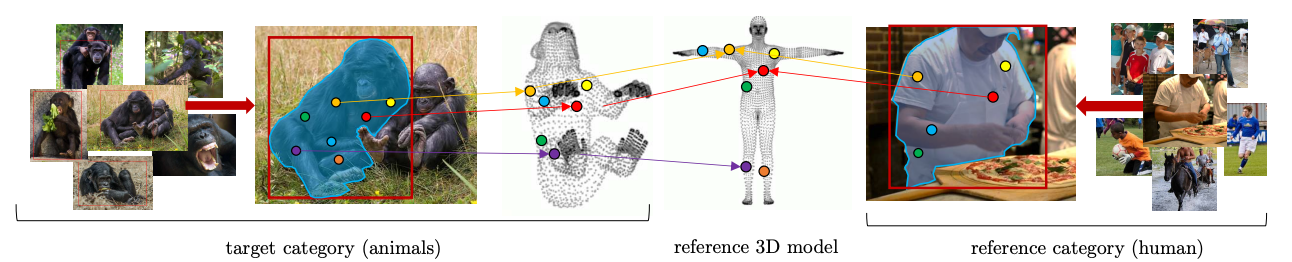}
\caption{Transfering human pose estimation knowledge to proximal animal classes (\eg Chimpanzees) for their pose estimation \citep{DensePose_Chimpanzees}.}
\label{Fig_human_to_chimp_APE}
\end{figure*}

Exploring APE through various sensory and modality inputs is important as animals may reside in diverse habitats which require different sensors for data collection. Most current APE methodologies are RGB-based, which predict the articulated joint positions from images or videos. However, RGB-based methods fail in many scenarios, such as nocturnal monitoring. For example, rodents are more active in the dark, and standard RGB cameras become inadequate due to poor lighting conditions. Consequently, infrared sensors have emerged as suitable tools for capturing the nocturnal activities of animals \citep{wang2018automated, patel2023animal}. Additionally, mmWave radar can detect movements but suffers from the scarcity of its generated point clouds; LiDAR adds value by providing depth information. Inertial sensors \citep{patel2017tracking} can also track movement and orientation, offering invaluable insights but requiring on-animal attachment. In specific scenarios, even auditory cues \citep{li2022sound} can indicate animal movement. This paper also explores the potential of using multiple sensors (\ie RGB + acoustic) to improve APE performance.

There are two representations for the final estimated animal poses regardless of our sensors or modalities of the APE input data. The first encompasses either two-dimensional (2D) or three-dimensional (3D) keypoint pose representations \citep{karashchuk2021anipose, li2023scarcenet, Nath2019, OpenApePose_2023} (\Cref{fig:Skeleton Pose Representations}). In this category, the animal's anatomy is represented as a tree of interconnected keypoints. These keypoints present animal body joints, and their interconnections reflect the natural relationship between these joints, providing a skeletal representation of the animal's posture. The second method, body mesh reconstruction \citep{zuffi20173d, ruegg2022barc, wu2023magicpony, yang2022banmo, wu2023dove}, presents a 3D view of the animal’s morphology (\Cref{fig:Body Mesh Reconstructions}). Within this framework, we distinguish between model-based and model-free reconstructions. The model-based technique (\eg SMAL \citep{zuffi20173d}) uses supervision from the predefined template shape of the creature learned from 3D toy animal scans. This template is then refined by optimizing the shape and pose parameters. In particular, the shape parameters capture variations in body metrics (\eg proportions, height, and weight), while the pose parameters address observable deformations due to movement. The model-based approach can generate detailed models, but collecting the necessary data for new types of animals can be time-intensive, restricting their applicability to rare species. In contrast, the model-free approach operates without any supervision from the predefined template shape of the creature (\eg MagicPony \citep{wu2023magicpony}, DOVE \citep{wu2023dove} and 3D-Fauna \citep{li2024learning}), which are more capable when estimating the animal poses for the diverse animal species.

The transition from HPE to APE is based on HPE frameworks. Due to skeletal similarities between humans and certain animals like quadrupeds, HPE models for skeletal keypoint detection \citep{wang2020deep, DeeperCut_Eldar_2016, deb2023keypoint} and body mesh reconstruction \citep{loper2023smpl} can be pertinent to APE, with modifications to cater to distinct anatomical and dynamic variations across animal species \citep{li2023scarcenet, rashid2017interspecies, zuffi20173d, ruegg2022barc}. One notable example is \citep{DensePose_Chimpanzees}, which involves transferring human DensePose knowledge to proximal animal classes (\Cref{Fig_human_to_chimp_APE}). This approach leverages the DensePose-COCO dataset, which contains dense human body annotations, and adapts it to animal species like chimpanzees. The transfer is facilitated by establishing correspondences between the 3D models of humans and animals using semantic alignment, so that the learned dense pose mappings for humans can be applied to the animal class. This enables pose extraction without dense animal-specific annotations. In addition, innovations in APE that allow adaptation to unseen animals and environmental changes can also improve the robustness of HPE, particularly in challenging scenarios like emergency response, where the surrounding environment changes rapidly. However, bridging these two fields requires adaptation of knowledge and consideration of the distinct challenges and ethical issues in human and animal studies. Ethical and responsible practices \citep{responseAI} are especially critical in HPE methods development, and similar approaches can be applied to the APE methods as well.

\subsection{Challenges in Human and Animal Pose Estimation}

HPE and APE encounter several common challenges that can impact the accuracy of associated algorithms. For example, humans and animals may appear differently due to their age, body composition, and clothing. Furthermore, we can see that a subject's perceived pose, whether human or animal, can significantly alter depending on the viewpoint. Another shared challenge is occlusion. Self-occlusions always exist, and certain body parts may be obscured by external objects (\eg cars for humans, trees for wild animals). Environmental factors can be another factor. For example, on a rainy day, it is harder to identify the humans or animals' poses in an outdoor scenario, this further introduces ambiguities. Lastly, a busy or cluttered background can make separating the subject from its environment difficult.

There are also unique challenges in APE. The animal kingdom has a much greater morphological diversity and inter/intra-species variations than humans. Different animal species exhibit significant domain discrepancies in physical characteristics, body structures, and postures. Even within the same species, there can be a large variety in body size, shape, texture, motion, and posture, presenting difficulties in creating universally applicable models \citep{ruegg2022barc}. For example, the pelage of porcupines can vary a lot, and this is often more pronounced and thicker in animals than in humans' hair, which can make tracking skeleton segments tricky. Furthermore, many animals exhibit natural camouflage, travel in herds, interact frequently, or share similar appearances within their species \citep{zuffi2019three}. These characteristics make it hard to distinguish individual animals and accurately estimate their poses as they blend into the background or appear similar to one another.

Environmental factors also play a pronounced role. Vegetation or terrain features in the natural environments may obstruct the view of animals. One example can be found in the AcinoSet dataset \citep{joska2021acinoset}, where the cheetahs' paws were excluded as the grass occluded them in most of the videos, making it difficult to estimate animal poses accurately. Moreover, changing lighting conditions can also heavily affect an animal's appearance, especially when the sun's position is moving in outdoor environments.

There are also many observational constraints in APE. For example, we often need to observe wildlife from afar; if we approach animals too closely, it may disturb their natural behaviour. This distant viewing distance may lead to reduced resolution and detail. In addition, animals can exhibit rapid and unpredictable movements (\eg the high-speed pursuit of a cheetah); thus, we may need a stable camera platform and high-speed capture techniques to avoid camera shake and motion blur. Also, nocturnal species may require specialized sensors like infrared or thermal sensors to capture their movements and poses.

Moreover, there are very limited annotated APE datasets \citep{yu2021ap}. Unlike HPE, which predominantly leans on supervised \citep{wang2020deep, cheng2020higherhrnet} and semi-supervised learning methodologies \citep{ukita2018semi, pavllo20193d, mitra2020multiview} sustained by large-scale human pose datasets such as MSCOCO \citep{lin2014microsoft} and MPII \citep{andriluka20142d}, APE is hindered by a lack of such datasets, especially for endangered species and wildlife \citep{zuffi2019three}. Also, it is often impractical to place animals in controlled environments to obtain ground truth poses, although zoos provide a semi-structured environment towards data collection. Therefore, unsupervised approaches that do not rely on labelled data have become popular in APE \citep{wu2023magicpony, li2021synthetic}.

\subsection{Applications of Animal Pose Estimation}

HPE has gained significant attention recently due to its applications in many domains, such as human-computer interaction, sports analytics, and healthcare. On the other hand, APE research has not been as prominent due to limited funding and methods. Despite this, APE is critically essential in various applications, as explained below and summarised in \Cref{APE_Application}.

\textbf{Tracking:}
APE has been applied in various tracking contexts; for example, in the CATER \citep{tracking_CATER_2023} framework, APE is combined with environment reconstruction to track ants in complex natural environments, providing fine-scale motion trajectories even in visually challenging conditions. The SMART-BARN \citep{Tracking_SMART_BARN_2023} system uses APE to track multiple animal species simultaneously in a large 3D arena and achieve accurate tracking and behavioural analysis in indoor and naturalistic environments. In the FaceMap \citep{tracking_2024facemap} framework, APE is used for tracking the orofacial movements of mice, enabling detailed analysis of their facial expressions and related neural activities. For honey bees, APE tracked an entire colony over extended periods, capturing complex social behaviours and interactions without needing physical tags, enhancing our understanding of colony dynamics and individual behaviours \citep{Honey_bee_tracking_2021}.

\begin{table}[!t]
\caption{Table listing various applications of animal pose estimation. The paper citation is available in column Year. \label{APE_Application}}
\centering
\setlength{\tabcolsep}{1.5pt}
\begin{tabular}{|c|c|l|l|l|} 
\hline
\textbf{Sl} & \textbf{Year} & \textbf{APE Paper} & \textbf{Animal Classes} & \textbf{Domain} \\
\hline

  1 & \citeyear{Bain_2019_ICCV_Chimp} & Bain \etal & Chimpanzee & Tracking \\
  
  2 & \citeyear{Honey_bee_tracking_2021} &  Bozek \etal & Honey Bee & Tracking \\ 

  3 & \citeyear{I_MuPPET_2022} & I-MuPPET  & Pigeon & Tracking \\
  
  4 & \citeyear{tracking_CATER_2023} & CATER  & Desert Ants & Tracking \\
  
  5 & \citeyear{Tracking_SMART_BARN_2023} & SMART-BARN  & Insects, Mammals, & Tracking \\
     &  &  & Birds  &  \\
  6 & \citeyear{Naik_2023_CVPR} & 3D-POP  & Birds & Tracking \\
  
  7 & \citeyear{wiltshire2023deepwild} & DeepWild  & Chimpanzees,   & Tracking  \\
       &  &  & Bonobos &  \\
       
  8 & \citeyear{tracking_2024facemap} & Facemap  & Mice & Tracking  \\
  
  9 & \citeyear{3D_MuPPET_2024} & 3D-MuPPET  & Pigeon & Tracking \\
  \hline
  
  10 & \citeyear{articulation_2023} & Stathopoulos \etal  & Sheep, Cows, Bear, & Articulation \\
  & &  & Giraffe, Horses & \\
  \hline
  
  11 & \citeyear{mouse_behav_analysis_2022} & Sheppard \etal  & Mouse & Behaviour \\
  
  12 & \citeyear{macaque_behaviour_analysis_2023} & Voloh \etal  & Macaque & Behaviour \\
  
  13 & \citeyear{macaques_Behavior_2023} & BARN  & Macaque  & Behaviour \\
  
  14 & \citeyear{gelada_monkey_behaviour_2023} & Koger \etal  & Gelada Monkeys & Behaviour \\
  
  15 & \citeyear{AmadeusGPT_2023} & AmadeusGPT  & Mammals & Behaviour \\
  
  16 & \citeyear{testard2024neural} & Testard \etal  & Macaque & Behaviour \\
  \hline
\end{tabular}
\end{table}

\textbf{Articulation:}
The authors of \citep{articulation_2023} explore how APE can be used to understand animal articulation by predicting their 3D shapes from monocular images. They demonstrate that training a 2D keypoint estimator on a small set of labelled images (50-150) can generate pseudo-labels on a more extensive set of unlabeled web images. Then, these pseudo-labels are used to train models for 3D shape reconstruction of various articulated quadrupeds.

\textbf{Behaviour Analysis:}
In agriculture, animal behaviour analysis based on APE plays an important role by enabling effective livestock management and health surveillance, which in turn provides continuous monitoring of movement and also early detection of animal's illness, thereby ensuring optimal animal welfare \citep{fang2021pose, russello2022t, farahnakian2021multi, gong2022multicow, doornweerd2021across, fang2022study}. Recently, more APE methods have been applied to study the behaviour of more species, such as macaques \citep{macaque_behaviour_analysis_2023, macaques_Behavior_2023, testard2024neural}, Gelada monkeys \citep{gelada_monkey_behaviour_2023}, and mice \citep{mouse_behav_analysis_2022}. Since 2023, there has been a notable shift towards integrating language prompts with APE in behaviour analysis research, exemplified by developing models like AmadeusGPT \citep{AmadeusGPT_2023}. This combination has opened up new possibilities for more comprehensive interpretations of animal behaviour.

\subsection{Scope of this Survey}

We aim to provide a systematic review of APE, focusing on developments in deep learning-based methodologies for both 2D and 3D scenarios. It categorizes existing APE methods according to their sensor and modality inputs rather than limiting the scope to monocular images or videos.

Several existing reviews have been reported on HPE and APE separately. Prior surveys \citep{zheng2020deep, ji2009advances, sarafianos20163d, khan2018review, gong2016human, liu2022recent, song2021human, dubey2023comprehensive, wang2021deep, moeslund2001survey, moeslund2006survey, poppe2007vision, zhu2023human} have primarily concentrated on vision-based human motion capture, encompassing aspects of pose estimation, motion tracking, action recognition and motion generation. While these studies have significantly contributed to HPE, there remains a conspicuous absence of in-depth comparison and analysis between HPE and APE techniques. Reference \citep{jiang2022animal} discusses vision-based APE between 2013 and 2021, yet it falls short of offering an exhaustive discussion on the existing challenges and future directions in APE. Another limitation is that only a limited number of reviews \citep{zheng2020deep} discuss HPE or APE from sources other than vision-based inputs. This paper aims to bridge these gaps in the literature by offering a detailed analysis that extends beyond the scope of prior surveys and reviews:

\begin{itemize}
    \item We meticulously reviewed 176 papers up to 2024, categorizing them based on different parameters such as their output forms (\eg 2D \vs 3D pose estimation; keypoint \vs mesh pose representation), input sensor and modality types (monocular inputs \vs other sensory inputs; uni-modal APE \vs multi-modal APE), learning paradigms (\eg heatmap-based methods \vs regression methods; top-down methods \vs bottom-up methods; supervised \vs semi-supervised \vs self-supervised \vs unsupervised APE methods), and experimental setup (\eg single-camera-view \vs multiple-camera-views approaches; indoor \vs in-the-wild setup; the range of the sensors).

    \item We summarised and discussed datasets and evaluation metrics for both 2D and 3D APEs not only based on RGB cameras but also include other sensors and modalities such as motion capture (MoCap) systems, RGBD, LiDAR, infrared, 3D Scanner, IMU, acoustic and language cues.

    \item We delved into the potential and limitations of current multi-sensor and multi-modal APE, highlighting how these approaches can complement or enhance unimodal APE methodologies.
    
    \item We presented an extensive comparison between HPE and APE. This includes the transferability of knowledge and techniques between these two fields and how innovations in APE can reciprocally enrich HPE and the broader machine learning paradigm. It also explained the trend towards self-supervised and unsupervised methods for APE and compared the advantages and disadvantages of the existing supervised methods.
    
    \item We also provided an overview of a wide array of applications of APE, ranging from wildlife conservation, biomechanics and neuroscience to 3D animal asset articulation, illustrating the practical implications and the interdisciplinary nature of APE.
    
    \item Finally, we thoroughly discussed the key challenges in APE and pointed towards potential future research avenues to enhance performance and applicability.
\end{itemize}

\section{Datasets} \label{datasetsection}

\begin{table*}[htbp] 
\renewcommand{\arraystretch}{1.0}
\providecommand{\symbyes}{{\includegraphics[width=0.0125\textwidth]{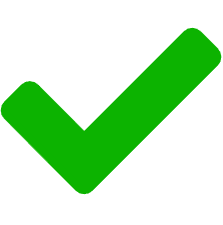}}}

\providecommand{\symbno}{{\includegraphics[width=0.0125\textwidth]{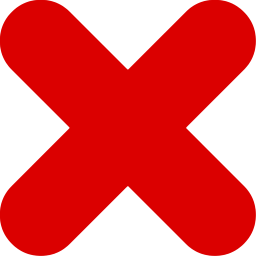}}} 
\providecommand{\symbarticulation}{{\includegraphics[width=0.0125\textwidth]{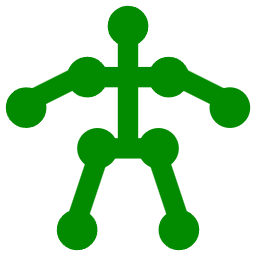}}} 
\providecommand{\symbshape}{{\includegraphics[width=0.0125\textwidth]{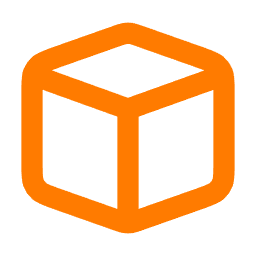}}} 
\providecommand{\symbappearance}{{\includegraphics[width=0.0125\textwidth]{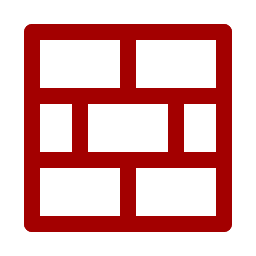}}}
\providecommand{\symbflow}{{\includegraphics[width=0.0125\textwidth]{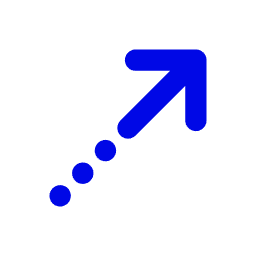}}}
\providecommand{\symbcorrespondence}{{\includegraphics[width=0.0125\textwidth]{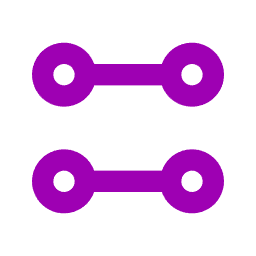}}}
\providecommand{\symbimage}{{\includegraphics[width=0.0125\textwidth]{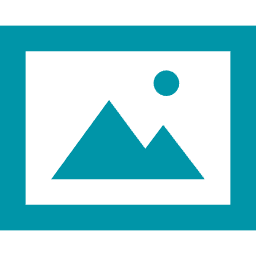}}}
\providecommand{\symbmultiview}{{\includegraphics[width=0.0125\textwidth]{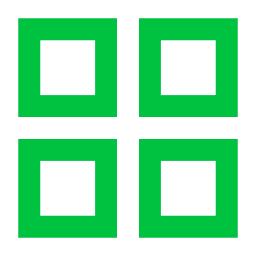}}}
\providecommand{\symbvideo}{{\includegraphics[width=0.0125\textwidth]{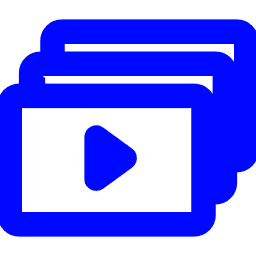}}}
\providecommand{\symbdepth}{{\includegraphics[width=0.0125\textwidth]{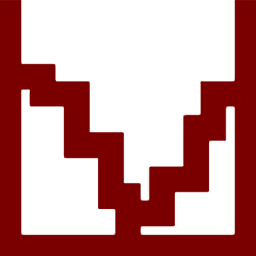}}}
\providecommand{\symbautoencoder}{{\includegraphics[width=0.0125\textwidth]{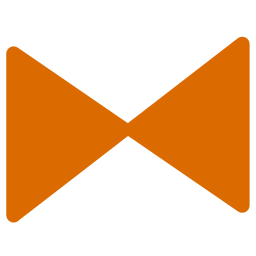}}}
\providecommand{\symbautodecoder}{{\includegraphics[width=0.0125\textwidth]{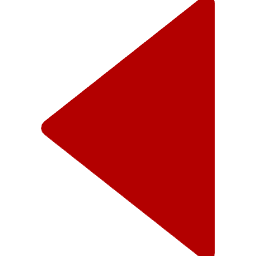}}}
\providecommand{\symbgan}{{\includegraphics[width=0.0125\textwidth]{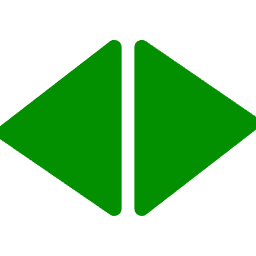}}}
\providecommand{\symbdiffusion}{{\includegraphics[width=0.0125\textwidth]{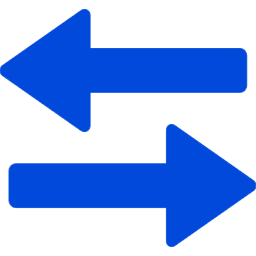}}}

\providecommand{\symbmultiviewvideo}{\includegraphics[trim={0 1.5cm 0 0},clip,width=0.0145\textwidth]{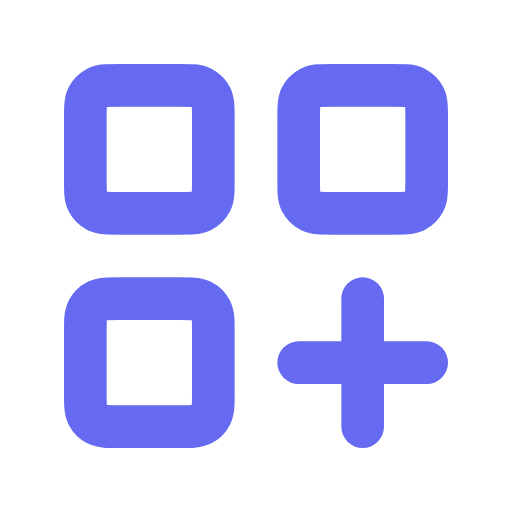}}
\providecommand{\symbmultivideo}{\scalerel*{\includegraphics{multiple-screen-play-svgrepo-com.png}}{B}}
\providecommand{\symbevent}{\includegraphics[width=0.0125\textwidth]{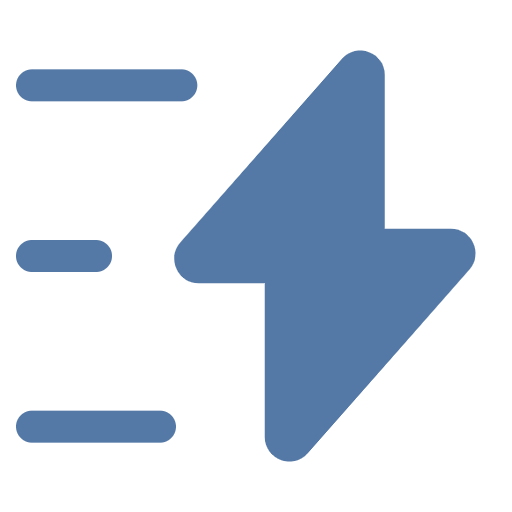}}
\providecommand{\symblidar}{\includegraphics[width=0.0125\textwidth]{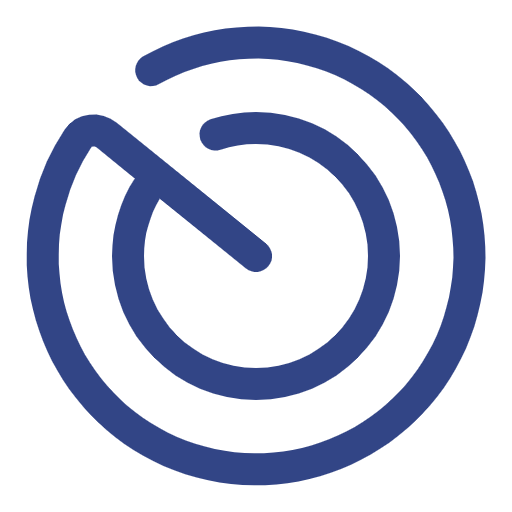}}
\providecommand{\symbmesh}{\includegraphics[trim={0 1cm 0 0},clip,width=0.015\textwidth]{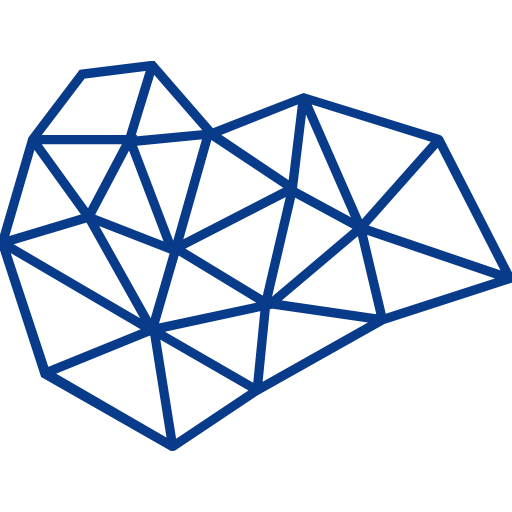}}
\providecommand{\symbef}{\includegraphics[width=0.0125\textwidth]{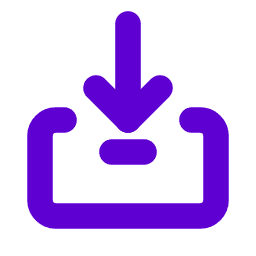}}
\providecommand{\symbmultiscene}{\includegraphics[width=0.0125\textwidth]{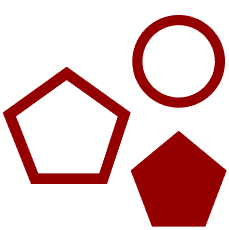}}
\providecommand{\symbif}{\includegraphics[width=0.0125\textwidth]{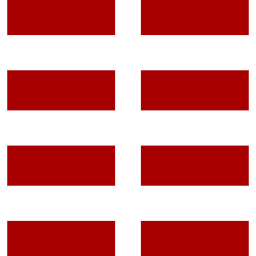}}

\providecommand{\symbmask}{\includegraphics[width=0.0125\textwidth]{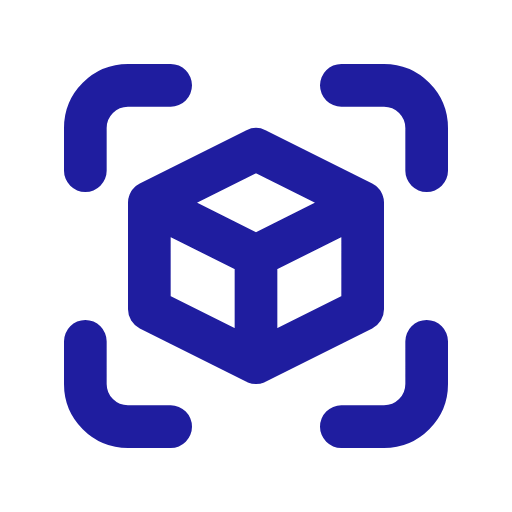}}
\providecommand{\symbsemanticsegmentation}{\includegraphics[width=0.0125\textwidth]{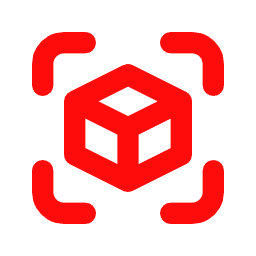}}
\providecommand{\symbof}{\includegraphics[trim={0 4cm 0 0},clip,width=0.02\textwidth]{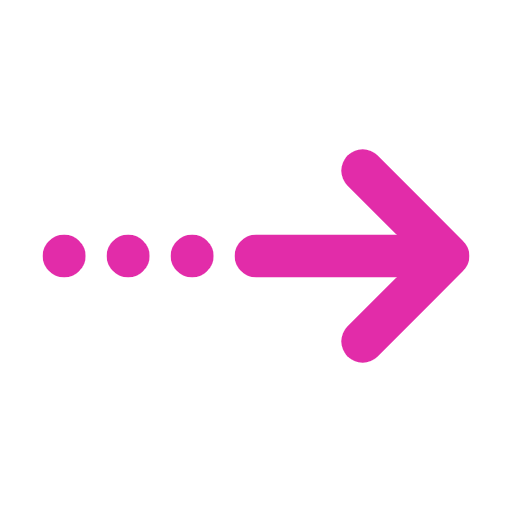}}
\providecommand{\symbpd}{\includegraphics[width=0.0125\textwidth]{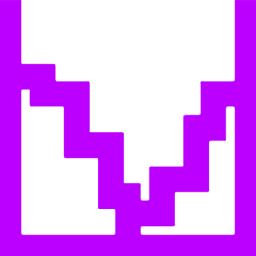}}
\providecommand{\symbdp}{\includegraphics[width=0.0125\textwidth]{arrows-svgrepo-com.png}}

\providecommand{\symbmlp}{\includegraphics[width=0.0115\textwidth]{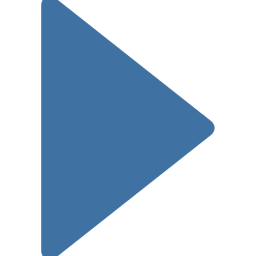}}
\providecommand{\symbtransformer}{\includegraphics[width=0.0115\textwidth]{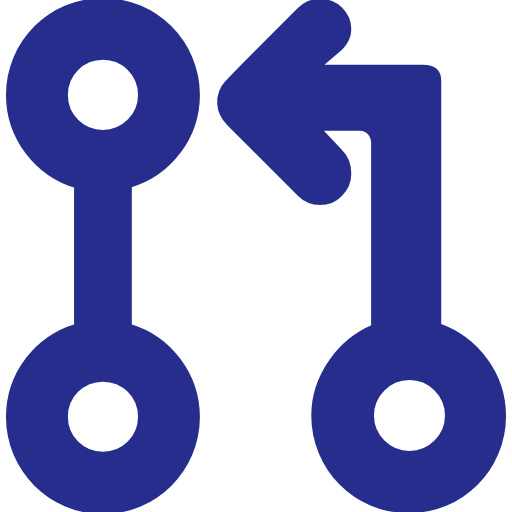}}
\providecommand{\symbpc}{\includegraphics[width=0.0125\textwidth]{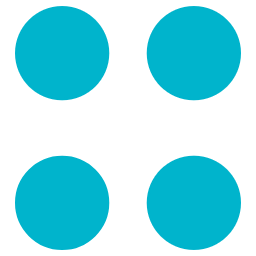}}
\providecommand{\symbvoxel}{\includegraphics[width=0.0125\textwidth]{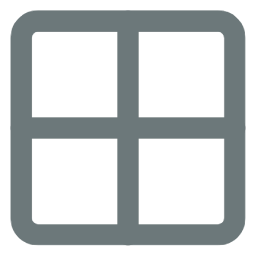}}
\providecommand{\symbtriplane}{\includegraphics[trim={0 0.5cm 0 0},clip,width=0.0135\textwidth]{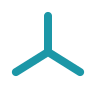}}
\providecommand{\symbtensorfac}{\includegraphics[trim={0 0.5cm 0 0},clip,width=0.0135\textwidth]{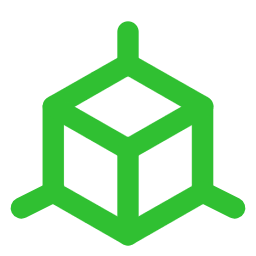}}
\providecommand{\symboctree}{\includegraphics[width=0.0125\textwidth]{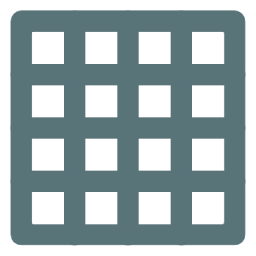}}
\providecommand{\symbmpi}{\includegraphics[width=0.0125\textwidth]{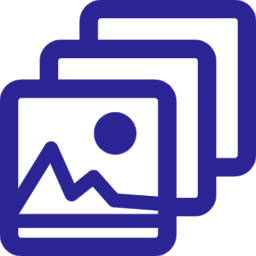}}
\providecommand{\symbibr}{\includegraphics[width=0.0125\textwidth]{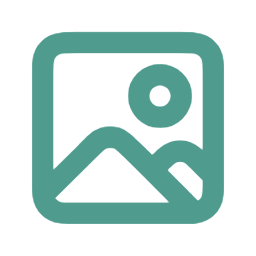}}

\providecommand{\symbdensity}{\includegraphics[width=0.0125\textwidth]{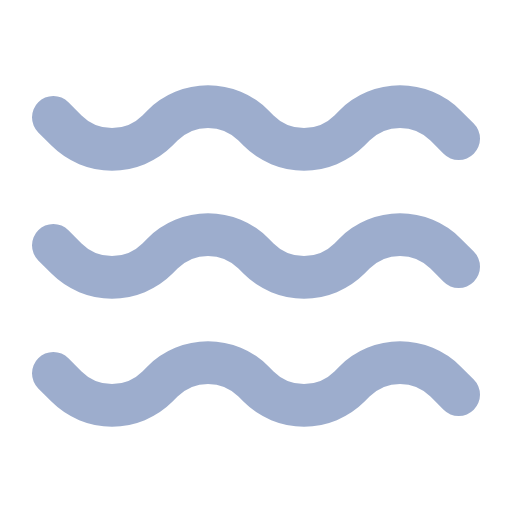}}
\providecommand{\symboccupancy}{\includegraphics[trim={1cm 1.5cm 0 0},clip,width=0.0125\textwidth]{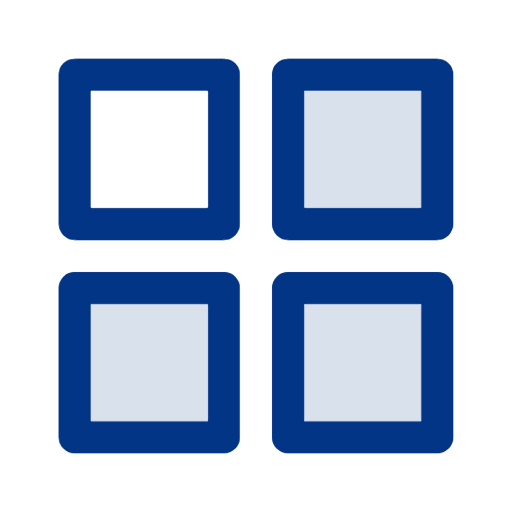}}
\providecommand{\symbsdf}{\includegraphics[trim={0 1cm 0 0},clip,width=0.0125\textwidth]{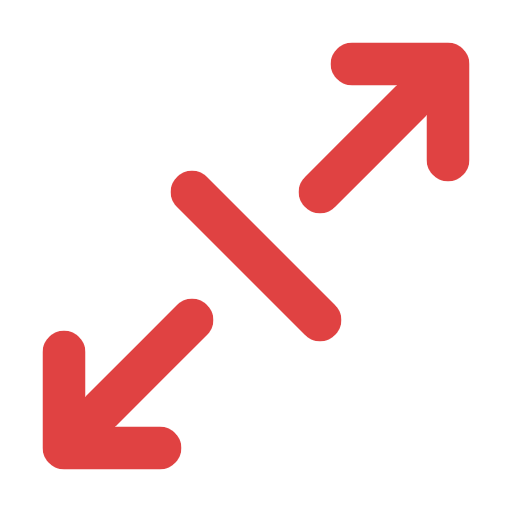}}
\providecommand{\symbradiance}{\includegraphics[width=0.0125\textwidth]{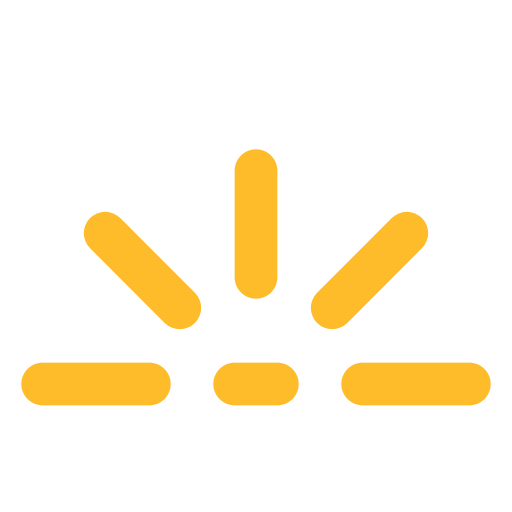}}
\providecommand{\symbgaussians}{\includegraphics[width=0.0125\textwidth]{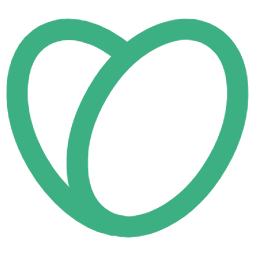}}
\providecommand{\symbsurfels}{\includegraphics[width=0.0125\textwidth]{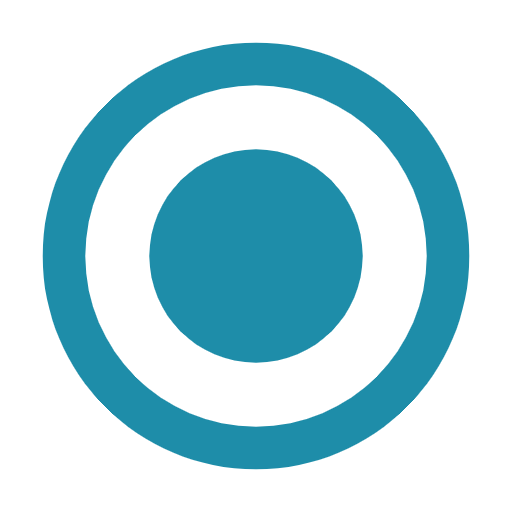}}
\providecommand{\symbsemantics}{\includegraphics[width=0.0125\textwidth]{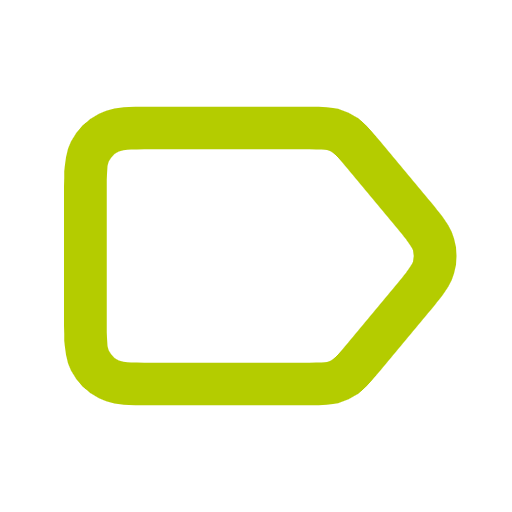}}

\providecommand{\symbphy}{\includegraphics[width=0.0125\textwidth]{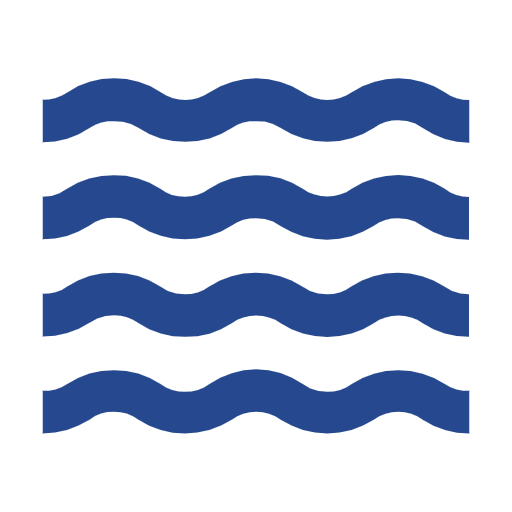}}
\providecommand{\symbposition}{\includegraphics[trim={0 -4cm 0 0},clip,width=0.0075\textwidth]{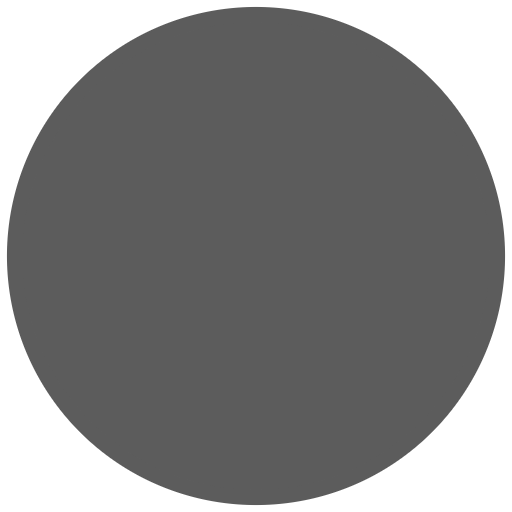}}
\providecommand{\symbdisplacement}{\includegraphics[trim={2cm 1.5cm 0 0},width=0.0125\textwidth]{flow-svgrepo-com.png}}
\providecommand{\symbrigid}{\includegraphics[width=0.0125\textwidth]{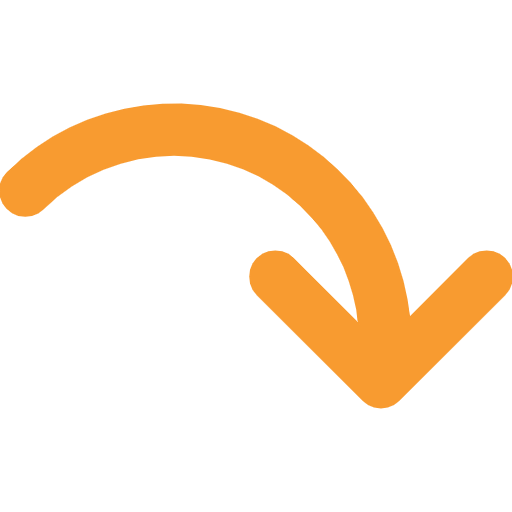}}
\providecommand{\symbaffine}{\symbrigid\,\includegraphics[width=0.0125\textwidth]{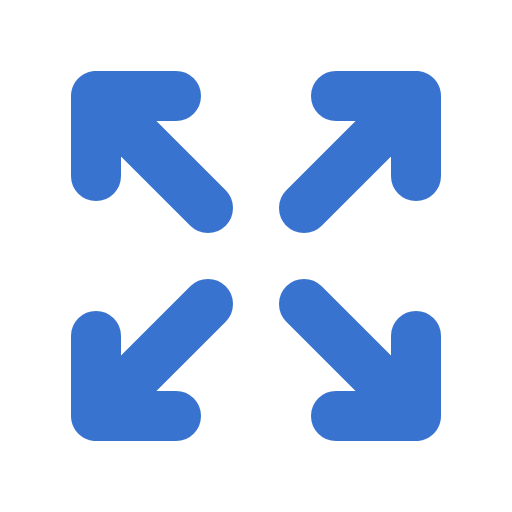}}
\providecommand{\symbvelocityfield}{\includegraphics[trim={2cm 0 0 0},clip,width=0.0125\textwidth]{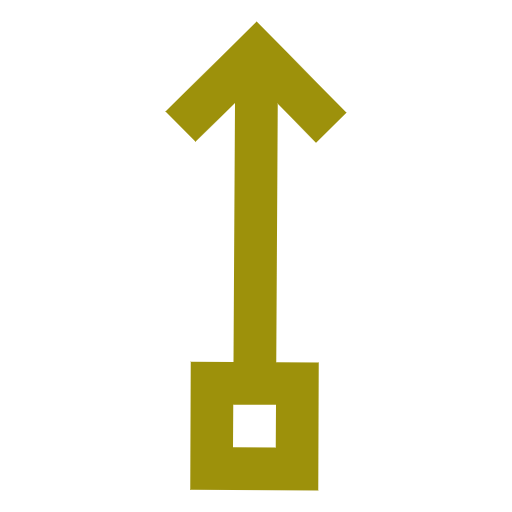}}
\providecommand{\symbmotiontrajectory}{\includegraphics[trim={6cm 1.25cm 0 0}, clip, width=0.01\textwidth]{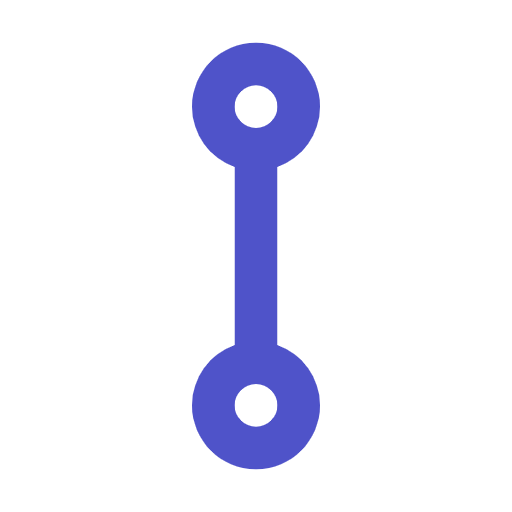}}
\providecommand{\symbjoint}{\includegraphics[width=0.0125\textwidth]{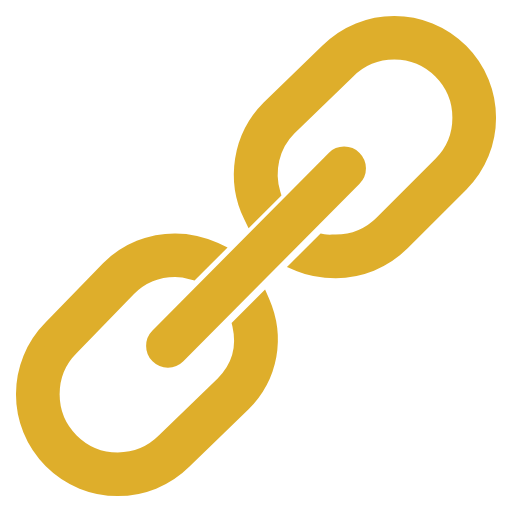}}
\providecommand{\symbvb}{\includegraphics[width=0.0125\textwidth]{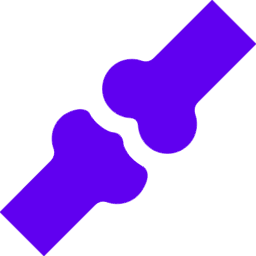}}
\providecommand{\symbskel}{\includegraphics[width=0.0125\textwidth]{person-skeleton-svgrepo-com.png}}
\providecommand{\symbcage}{\includegraphics[width=0.0125\textwidth]{cube-svgrepo-com.png}}
\providecommand{\symbedg}{\includegraphics[width=0.0125\textwidth]{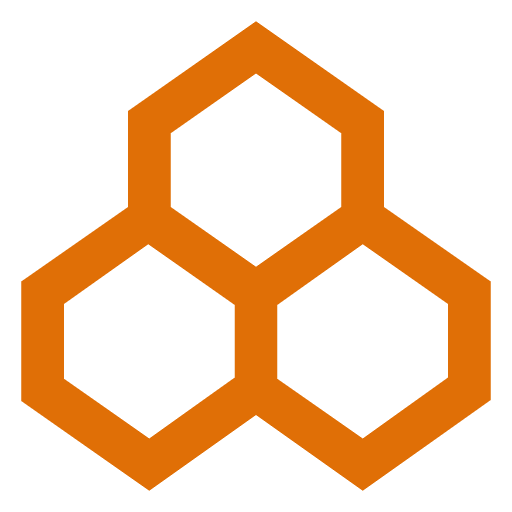}}
\providecommand{\symblbs}{{\includegraphics[width=0.0125\textwidth]{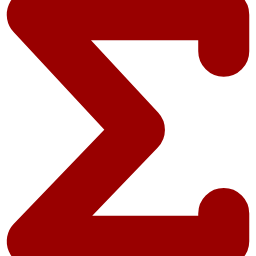}}
}
\providecommand{\symbdqb}{{\includegraphics[width=0.0125\textwidth]{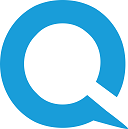}}}
\providecommand{\symbfdm}{{\includegraphics[width=0.0125\textwidth]{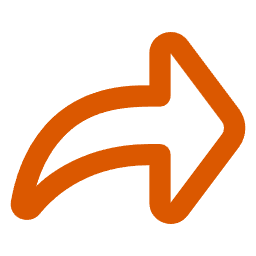}}}
\providecommand{\symbscrew}{{\includegraphics[width=0.0125\textwidth]{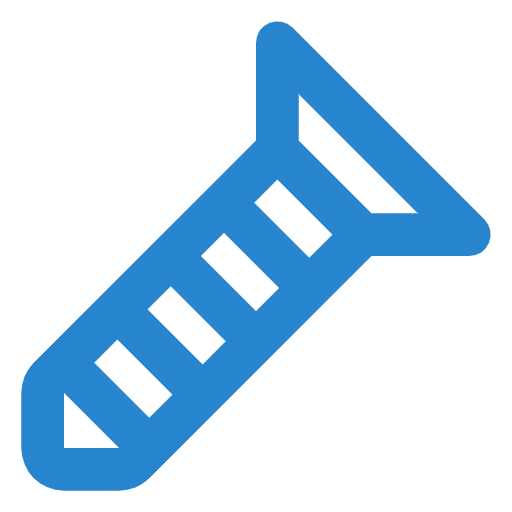}}}

\providecommand{\symbfd}{{\includegraphics[width=0.0125\textwidth]{forward-svgrepo-com.png}}}
\providecommand{\symbbd}{{\includegraphics[width=0.0125\textwidth]{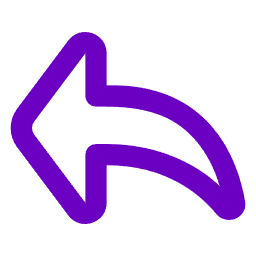}}}
\providecommand{\symbbid}{{\includegraphics[width=0.0125\textwidth]{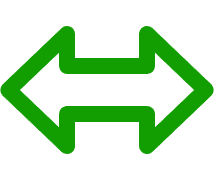}}}

\providecommand{\symbcm}{\includegraphics[width=0.014\textwidth]{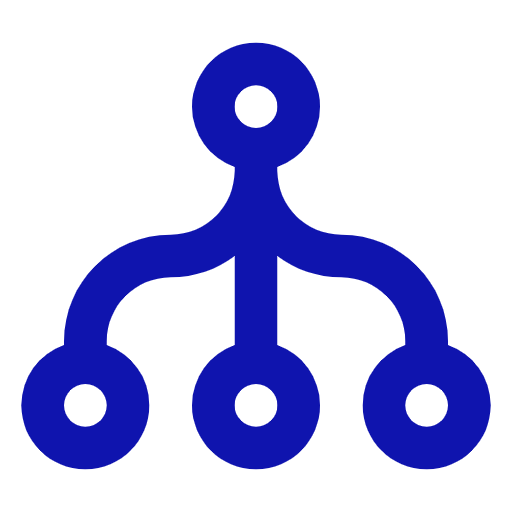}}
\providecommand{\symbifm}{\includegraphics[width=0.014\textwidth]{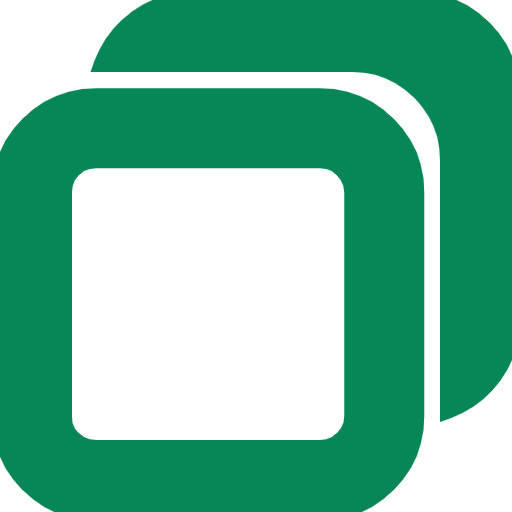}}
\providecommand{\symbdb}{{\includegraphics[width=0.014\textwidth]{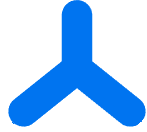}}}
\providecommand{\symbftf}{\includegraphics[width=0.014\textwidth]{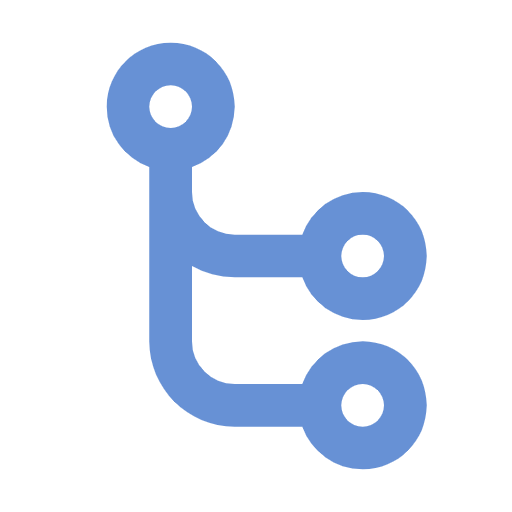}}

\providecommand{\symboff}{\includegraphics[width=0.0125\textwidth]{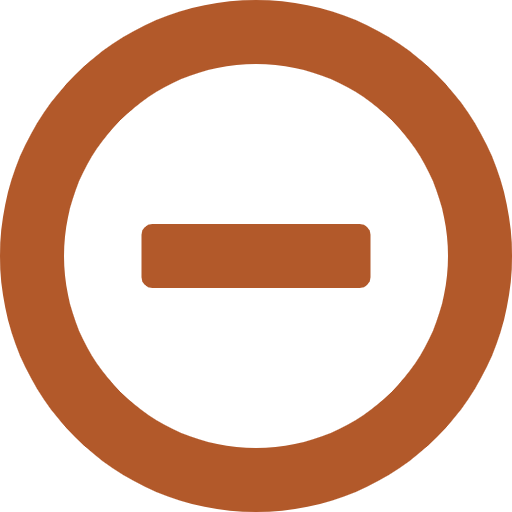}}
\providecommand{\symbon}{\includegraphics[width=0.0125\textwidth]{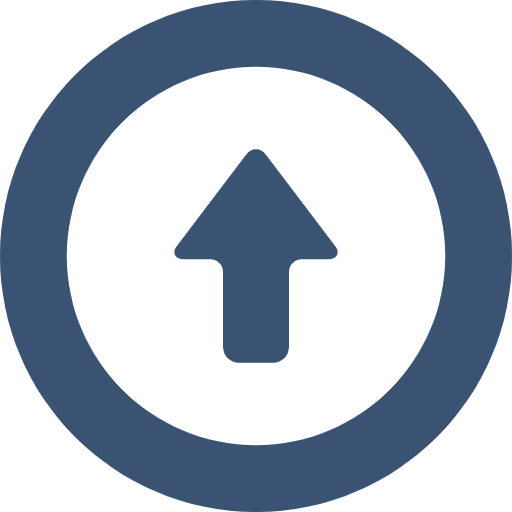}}
\providecommand{\symbrr}{\includegraphics[width=0.014\textwidth]{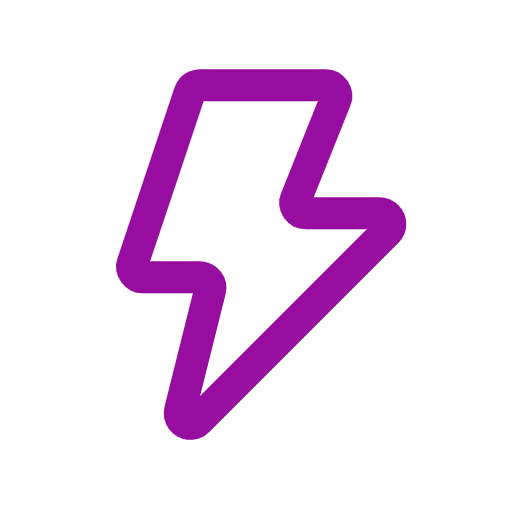}}
\providecommand{\symbreal}{\includegraphics[width=0.014\textwidth]{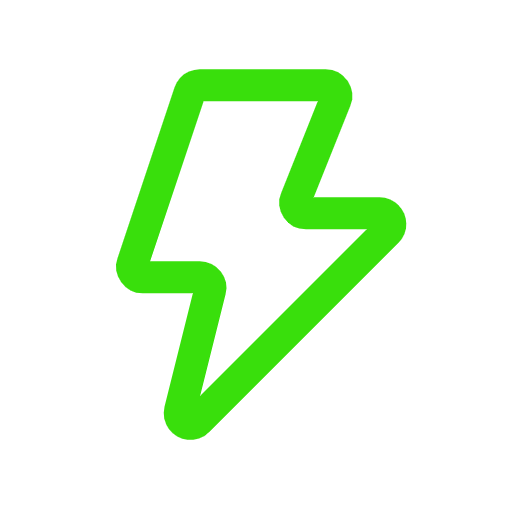}}

\providecommand{\symbcamera}{\includegraphics[width=0.014\textwidth]{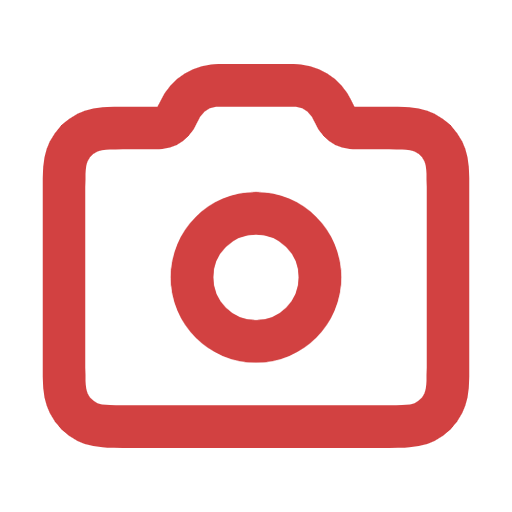}}
\providecommand{\symbobject}{\includegraphics[width=0.014\textwidth]{cube-svgrepo-com.png}}

\providecommand{\symbnf}{
\includegraphics[width=0.0125\textwidth]{eight-rectangles-divided-in-two-columns-svgrepo-com.png}
}
\providecommand{\symbbb}{{\includegraphics[width=0.0125\textwidth]{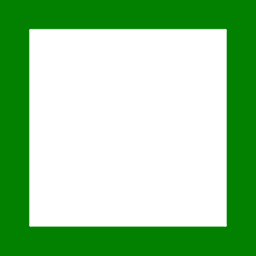}}}
\providecommand{\symbei}{
{\includegraphics[width=0.0125\textwidth]{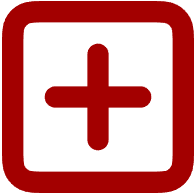}}
}
\providecommand{\symbsr}{
{\includegraphics[width=0.0125\textwidth]{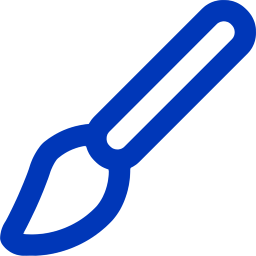}}
}
\providecommand{\symbla}{
\includegraphics[width=0.0125\textwidth]{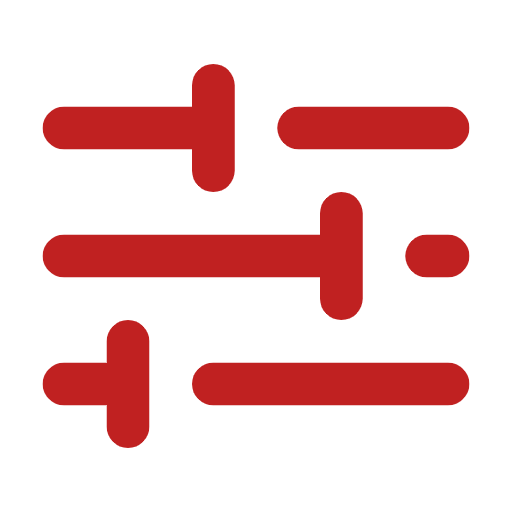}
}
\providecommand{\symbkp}{
{\includegraphics[width=0.0125\textwidth]{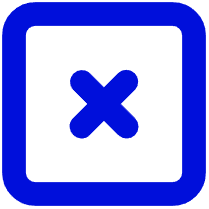}}
}
\providecommand{\symbact}{
{\includegraphics[width=0.0125\textwidth]{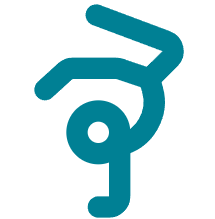}}
}

\providecommand{\symbds}{
\includegraphics[width=0.0125\textwidth]{multiple-screen-play-svgrepo-com.png}}

\providecommand{\symbcf}{
\includegraphics[width=0.0125\textwidth]{eight-rectangles-divided-in-two-columns-svgrepo-com.png}
}
\caption{RGB-based APE Dataset (Images-based). Column N indicates no. of animals: M denotes multiple animals' pose is annotated in one frame, while S denotes a single animal's pose is annotated.
In I/W column, $\symbsurfels$ and $\symbdqb$ represent indoor and in-the-wild setups, respectively. 
Exp. is Experiment.
The dataset citation is available in column Year.}
\small
\begin{center}
\setlength{\tabcolsep}{1.5pt}
\begin{tabular}{c|c|l|l|l|c|c|l} 
\hline
\textbf{Sl} & \textbf{Year} & \textbf{Datasets \& Size} & \textbf{Animal Classes} & \textbf{Source \& Exp. Set-up} &  \textbf{N} & \textbf{I/W} & \textbf{Annotation} \\
\hline
\midrule
  & \multicolumn{7}{c}{\textcolor{blue}{RGB-based APE Dataset - Images-based}} \\
 \midrule 
  1 & \citeyear{Stanford_dogs2011} & Stanford Dogs & Dogs (120 breeds) & a. ImageNet data & N/A & $\symbsurfels$ & Class label, bounding box  \\
  & & 20k images & & b. Single view & & & \\
  \hline
  
  2 & \citeyear{lin2014microsoft} & COCO  & Bear, Sheep, Zebra, & a. Online repositories & N/A & $\symbsurfels$ \& $\symbdqb$ & Bounding box, class label,\\
  & & 1k animals images & Cat, Dog, Horse, & b. Single view & & & segmentation \\
  & & & Giraffe &  & & &  \\
  \hline

  3 & \citeyear{alessio2018animals10} & Animals-10  & Dog, Horse, Sheep, & a. Google images & N/A & $\symbsurfels$ \& $\symbdqb$ & N/A\\
  & & 28k images & Cat, Cow, Elephant & b. Single view & &\\
  \hline
  
  4 & \citeyear{cao2019cross} & Animal Pose  & Dogs, Cats, Cows, & a. Animals-10 (\citeyear{alessio2018animals10}) \&  & S & $\symbsurfels$ \& $\symbdqb$ & 20 keypoints (2D pose), \\
  & & 4k images & Horses, Sheep \& & COCO (\citeyear{lin2014microsoft}) & & & bounding box\\
  & &  &  other 7 categories & b. Single view & & & \\
  \hline
  
  5 & \citeyear{zuffi2019three} & Grevy’s Zebra  & Synthetic Zebra & a. Computer generated  & S & $\symbdqb$ & 28 keypoints (2D pose),\\
  & & 12,850 images & & b. Single view & & & 3D mesh\\
  \hline
  
  6 & \citeyear{Mu_2020_CVPR} & Synthetic Animal  & Synthetic Hound, & a. COCO2017 (\citeyear{lin2014microsoft}) & S & $\symbsurfels$ \& $\symbdqb$ & 18 keypoints (2D\&3D pose), \\
  & & 10k images & Elephant, Tiger,  & b. Single view & & &  depth maps, segmentation\\
    & &  & Horse, Sheep &  & & & \\

  \hline

  7 & \citeyear {yu2021ap} & AP-10K  & 23 animal families, & a. Online repositories & M & $\symbdqb$ & 17 keypoints (2D pose)\\
  & & 10,015 images & 54 species & b. Single view  &  &  \\
  \hline

  8 & \citeyear{biggs2020wldo} & StanfordExtra  & Dogs& a. Stanford Dogs (\citeyear{Stanford_dogs2011}) & S & $\symbdqb$ & 24 keypoints (2D pose),\\
  & & 12k images & & b. Single view & & & segmentation\\
  \hline

  9 & \citeyear{MacaquePose_2021} & MacaquePose & Macaque Monkey & a. Manually collected \&  & M & $\symbdqb$ & 17 keypoint (2D pose) \\
  & & 13,000 images & & Online & & \\
    & &  &  & b. Single view & & & \\

  \hline
  
  10 & \citeyear{ng2022animal} & Animal Kingdom  & 850 animal species & a.  Online repositories & S & $\symbdqb$ & 23 keypoints (2D pose) \\
  & & 35k images & & b. Single view & & \\
  \hline
  
  11 & \citeyear{openmonkeychallenge_2022} & OpenMonkey  & 26 Monkey species & a. Manually collected \&  & S & $\symbdqb$ & 17 keypoints (2D pose)\\
  & & Challenge  & & Online & & \\
      & & 111k images &  & b. Single view & & & \\

  \hline
  
  12 & \citeyear{OpenApePose_2023} & OpenApePose  & 6 Ape species & a. Manually collected & S & $\symbdqb$ & 16 keypoints (2D pose)\\
  & & 71,868 images & & b. Single view  & & \\
  \hline
  
  13 & \citeyear{xu2024animal3d} & Animal3D  & 40 Mammal species &  a. PartImageNet (\citeyear{he2022partimagenet}) \&  & S & $\symbdqb$ & 26 keypoints (2D pose),\\
  & & 3379 images & & COCO (\citeyear{lin2014microsoft}) &  & & 3D mesh \\ 
    & &  &  & b. Single view & & & \\

\hline
\hline
\end{tabular}
\label{Dataset_APE_images}
\end{center}
\end{table*}

It is hard to obtain accurate ground truth poses for APE, especially for wild animals, due to the unpredictable natural environments, the wide range of animal behaviours, occlusions, and the lack of controlled setups for precise annotation. In this paper, we have categorized APE datasets into two types based on the data they provide and the sensor or modality they use: (1) RGB-based APE datasets, which rely only on RGB images or video sequences (often involving multiple frames) to provide animal pose annotations. (2) Other-sensor-based datasets, which utilize sensors and modalities from sources beyond RGB, such as MoCap systems, 3D scanners, infrared cameras, IMUs, and LiDAR to acquire pose information.

\subsection{RGB-based Animal Pose Estimation Dataset} \label{rgbapedataha}

In RGB-based APE datasets, we can further divide them into two categories based on the data type: 1) image-based APE datasets, which rely on individual RGB images to provide pose annotations for animals (\Cref{Dataset_APE_images}); 2) video-based APE datasets, which utilize video sequences, consisting of multiple frames, to capture animal pose labels over time (\Cref{Dataset_APE_2D_videos}).

\subsubsection{\textbf{Images-based Animal Pose Estimation Dataset}}

Early datasets in this category lacked large-scale annotated animal pose labels. For example, the Stanford Dogs (2011) dataset \citep{Stanford_dogs2011} includes 20k images of 120 dog breeds with class labels and bounding boxes but no pose keypoints. Similarly, the COCO (2014) dataset \citep{lin2014microsoft} with 1k images only provides annotations such as bounding boxes, segmentation, and class labels for animals like cats, dogs, and horses. The Animals-10 (2018) dataset \citep{alessio2018animals10}, consisting of 14k images of common animals such as dogs, cats, and horses, also lacks animal keypoint annotations.

As there are few pose-labeled animal datasets available, more 2D pose-labeled animal datasets have been developed to evaluate the performance of APE objectively. For instance, the Animal Pose (2019) \citep{cao2019cross} consists of 4k images across 10 animal categories, with each image annotated with 20 2D keypoints to represent the animal skeleton. The StanfordExtra (2020) \citep{biggs2020wldo} offers 12k dogs images with 24 2D pose keypoints. The AP-10K (2021) \citep{yu2021ap} provides 10k images from 23 animal families, with 17 2D keypoints labeled for each detected animal. The recent Animal Kingdom (2022) \citep{ng2022animal}, the OpenMonkeyChallenge (2022) \citep{openmonkeychallenge_2022}, and the OpenApePose (2023) \citep{OpenApePose_2023} all include annotated 2D animal poses keypoints.

Beyond 2D keypoint pose labels, more recent datasets have started providing 3D pose annotations, including 3D keypoints and meshes. Examples include the Synthetic Grevy’s Zebra (2019) \citep{zuffi2019three}, the Synthetic Animal (2020) \citep{Mu_2020_CVPR}, and the Animal3D (2023) \citep{xu2024animal3d}, which offer both high-quality 2D keypoint annotations and 3D pose labels. This development reflects the growing diversity and depth of datasets available for APE research.


\subsubsection{\textbf{Video-based Animal Pose Estimation Dataset}}

\begin{table*}[htbp]
\renewcommand{\arraystretch}{1.0}

\caption{RGB-based APE Dataset (Video-based). 
NGD denotes National Geographic Documentaries; the symbol $\symbcamera$ denotes cameras. 
Column $\symbcamera$ Rng is camera range. 
Exp. is Experiment.
The dataset citation is available in column Year.
All the other symbols and column meanings are defined in \Cref{Dataset_APE_images}.} 
\begin{center}
\small
\setlength{\tabcolsep}{1.5pt}
\begin{tabular}{c|c|l|l|l|c|c|c|l} 
\hline
\textbf{Sl} & \textbf{Year} & \textbf{Datasets \& Size} & \textbf{Animal Classes} & \textbf{Source \& Exp. Set-up} & \textbf{$\symbcamera$ Rng} & \textbf{N} & \textbf{I/W} & \textbf{Annotation} \\
\hline
 \midrule
 & \multicolumn{8}{c}{\textcolor{blue}{RGB-based APE Dataset - Videos-based}} \\
 \midrule
  
  14 & \citeyear{ delpero15cvpr} & TigDog  & Dogs, Horses, Tigers & a. YouTube \& NGD videos & N/A & S & $\symbdqb$ & 19 keypoints \\
  & & 110k frames, 55 videos &  & b. Single view & & & & (2D pose)  \\
  \hline
  
  15 & \citeyear{BADJA_biggs2018creatures} & BADJA  & Bears, Camels, Dogs, & a. DAVIS (\citeyear{DAVIS_video_data_2017}) videos & 1-20m & S & $\symbdqb$ & 20 keypoints \\
  & & 11 videos & Horses, Tigers, Cats, & b. Single view & & & & (2D\&3D pose), \\
    & &  & Cows  &  & & & & segmentation \\
  
  \hline


  16 & \citeyear{ATRW_Li_2020} & ATRW  & Amur tigers &  a. Manually collected & N/A & M & $\symbdqb$ & 15 keypoints  \\
  & & 8k videos & & b. Single view & & & & (2D pose) \\
  \hline

  17 & \citeyear{bala2020automated} & OpenMonkeyPose  & Monkeys & a. Manually collected  & 0-2.5m & M & $\symbsurfels$ & 13 keypoints  \\
  & &  195,228 frames & & b. Multi views (62 $\symbcamera$ views) & & & & (3D pose)\\
  \hline

  18 & \citeyear{joska2021acinoset} & AcinoSet  & Cheetahs 
  & a. Manually collected  & 3-12m & S & $\symbdqb$ & 20 keypoints  \\
  & & 119k frames, 93 videos & & b. Multi views (6 $\symbcamera$ views) & & & & (2D\&3D pose) \\
  \hline
  
  19 & \citeyear{mathis2021pretraining} & Horse-30  & Horses & a. Manually collected & N/A & S & $\symbdqb$ & 22 keypoints \\
  & & 8114 frames & & b. Single view & & & & (2D pose)\\
  \hline
  
  20 & \citeyear{yang2022apt} & APT-36K  & Quadruped Mammals& 
  a. YouTube videos & N/A & M & $\symbdqb$ & 17 keypoints \\
  & & 36k frames, 2.4k videos & (30 Species) & b. Single view & & & & (2D pose)\\
  
  \hline
  21 & \citeyear{chimpact_2023} & ChimpACT  & Chimpanzees & a. Manually collected & N/A & M & $\symbsurfels$ \& $\symbdqb$ & 16 keypoints  \\
  & & 16,028 frames, 163 videos & & b. Single view & & & & (2D pose) \\

\hline
\hline
\end{tabular}
\label{Dataset_APE_2D_videos}
\end{center}
\end{table*}

We have listed several well-known datasets in this category, with most pose annotations being presented in 2D and 3D keypoint formats (\Cref{Dataset_APE_2D_videos}). In particular, most 2D keypoint pose labels in these datasets (\eg Horse-30 \citep{mathis2021pretraining}, APT-36K \citep{yang2022apt}, ChimpACT \citep{chimpact_2023}) are provided by well-trained annotators. For instance, the TigDog (2016) \citep{delpero16ijcv, delpero15cvpr} includes 110k frames of in-the-wild dogs, horses, and tigers, with 19 keypoint annotations for 2D poses. The Horse-30 (2021) \citep{mathis2021pretraining} provides 8k frames of horses in-the-wild, with 22 annotated 2D keypoints. While these datasets are valuable, they are limited to individual animals. However, there is a recent growing trend toward providing multi-animal pose annotations (\eg ATRW \citep{ATRW_Li_2020}, APT-36K \citep{yang2022apt} and ChimpACT \citep{chimpact_2023}) to account for the frequent interactions between quadruped animals.

There is also a small collection of publicly available 3D animal pose benchmark datasets. For example, the BADJA (2018) \citep{BADJA_biggs2018creatures} contains 11 in-the-wild animal videos annotated with 20 3D joints mapped onto a SMAL mesh. The AcinoSet (2021) \citep{joska2021acinoset} offers 7k frames of hand-labelled 3D keypoints from cheetahs, capturing their running behaviour in-the-wild. Additionally, the OpenMonkeyPose (2020) \citep{bala2020automated} includes 195k well-annotated 3D keypoint pose instances, spanning a large variation of poses and positions of monkeys across diverse activities. It is worth noting that both the OpenMonkeyPose \citep{bala2020automated} and AcinoSet \citep{joska2021acinoset} include synchronized multiple-camera-views footage and camera calibration, enabling multiple-camera-views geometry and triangulation for precise 3D pose estimation (\eg OpenMonkeyPose offers 62-viewpoint, while AcinoSet uses 6). We also discuss using MoCap systems with markers to obtain ground truth 3D animal poses in the following subsection.

One common limitation across many datasets in this category is their focus on specific animal species. Due to the significant variations in appearance, behaviour, and joint distributions between species, APE models trained on these datasets often fail to generalize to unseen species, leading to poor performance in real-world scenarios. To fill this gap, the APT-36K (2022) \citep{yang2022apt} was introduced, covering 2k video clips of 30 quadruped species with high-quality 2D keypoint annotations. This enables inter-species generalization testing and reflects the need for models that work across diverse species.

\begin{table*}[htbp]
\renewcommand{\arraystretch}{0.8}

\caption{Other-Sensor-based APE Dataset. $\symbvb$ is markers; vol. is volume. 
The dataset citation is available in column Year.
All the other symbols and column meanings are defined in \Cref{Dataset_APE_images} and \Cref{Dataset_APE_2D_videos}.}

\begin{center}
\small
\setlength{\tabcolsep}{1.5pt}
\begin{tabular}{c|c|l|l|l|c|c|c|l} 
\hline
\textbf{Sl} & \textbf{Year} & \textbf{Datasets \& Size} & \textbf{Animal} & \textbf{Source/Experiment} & \textbf{$\symbcamera$ Rng} & \textbf{N} & \textbf{I/W} & \textbf{Annotation} \\
& & & \textbf{Classes} & \textbf{set-up} & & \\
\hline

\midrule
 & \multicolumn{8}{c}{\textcolor{blue}{Other-Sensor-based APE Dataset - RGB+MoCap-based}} \\
 \midrule

  22 & \citeyear{pairr2m} & PAIR-R24M  & Rat & a. Manually collected & N/A & M & $\symbsurfels$ & 12 keypoints \\
  & & 24.3M frames & & b. Multi views (6 RGB video $\symbcamera$) & & & & (3D pose) \\
  & &  & & c. 1 MoCap $\symbcamera$ \& 12 $\symbvb$  & & & &  \\
  \hline

  23 & \citeyear{dunn2021geometric} & Rat7M  & Rat & a. Manually collected & N/A & S & $\symbsurfels$ & 20 keypoints\\
  & & 7M frames & & b. Multi views (6 RGB video $\symbcamera$) & & & & (3D pose) \\
  & &  & & c. 12 MoCap $\symbcamera$ \& 20 $\symbvb$ & & &\\
  \hline

  24 & \citeyear{Naik_2023_CVPR} & 3D-POP  & Pigeons & a. Manually collected &  & M & $\symbsurfels$ & 9 keypoints \\
  & & 300k frames,  & & b. Multi views (4 RGB $\symbcamera$) & 0-4.2m (RGB)  & & & (2D \& 3D pose) \\
  & & 4 videos & & c. 30 MoCap $\symbcamera$ \& 8 $\symbvb$ & 15x7x4m  (MoCap) & &\\   

  \hline
  25 & \citeyear{li2024poses} & PFERD  & Horses & a. Manually collected &16x20m & S & $\symbsurfels$ & 3D Mesh, \\ 
  & & 55+ hrs video & & b. Multi views (10 RGB $\symbcamera$) & from the &  &  &  2D joints \\
  & & & & c. 56 MoCap $\symbcamera$ \& over 100 $\symbvb$ &  arena center & & &  \\
  
\midrule
 \textcolor{blue}{} & \multicolumn{8}{c}{\textcolor{blue}{Other-Sensor-based APE Dataset - RGB+Depth+MoCap-based}} \\
 \midrule

  26 & \citeyear{RGBD_Dogs_2020_CVPR} & RGBD-Dog  & Dogs & a. Manually collected & 0.5-4.5m& S & $\symbsurfels$ & 3D Mesh, \\ 
  & & No size given & & b. Multi views (6 RGBD sensors) & (Depth) & & &  3D Skeletal data,\\
  & & & & c. 20 infrared $\symbcamera$ \& MoCap suit & N/A (infrared)& & & RGBD video \\
  
\midrule
 \textcolor{blue}{} & \multicolumn{8}{c}{\textcolor{blue}{Other-Sensor-based APE Dataset - RGB+LiDAR-based}} \\
 \midrule
  27 & \citeyear{WildPose2024} & WildPose  & Lion, & a. Manually collected & 2–120m & M & $\symbdqb$ &  15 keypoints, \\
  & & No size given & Springbok & b. Single view & & & & 3D point cloud \\ 

\midrule
 \textcolor{blue}{} & \multicolumn{8}{c}{\textcolor{blue}{Other-Sensor-based APE Dataset - Infrared-based}} \\
 \midrule
 
 28 & \citeyear{LoTE_Animal_2023_ICCV} & LoTE-Animal  & 11 animal & a. Manually collected \& Online & N/A & S & $\symbdqb$ & 17 keypoints\\
  & & 33k images & species & b. Single view & & & & (2D pose)\\
  
 \midrule
 \textcolor{blue}{} & \multicolumn{8}{c}{\textcolor{blue}{Other-Sensor-based APE Dataset - 3D scanner-based}} \\
 \midrule 

  29 & \citeyear{zuffi2024varen} & VAREN  & Horses & a. Manually collected & N/A & S & $\symbsurfels$ & 3D Mesh \\ 
  & & 4000 3D scans &  & b. No. of scanner Not given & & &\\
 
  \midrule
  \textcolor{blue}{} & \multicolumn{8}{c}{\textcolor{blue}{Other-Sensor-based APE Dataset - MoCap+IMU-based}} \\
  \midrule 
  30 & \citeyear{yigit2022wearable} & Yigit \etal  & Horse & a. Manually collected & & S & $\symbsurfels$ \& $\symbdqb$ & 3D Motion data\\
  & & No size given & & b. Multi views & & &  \\
  & & & & c. MoCap system (8 $\symbcamera$ \& 37 $\symbvb$) & N/A (MoCap) & & \\
  & & & & d. 4 IMUs  &  0-50m (IMU) & & \\
  \hline
  \hline
\end{tabular}
\label{3D_Dataset_APE}
\end{center}
\end{table*}

\subsection{Other-Sensor-based Animal Pose Estimation Dataset}

With developments in sensor technology, more APE datasets start to incorporate data from sources beyond traditional RGB cameras, such as depth sensors, MoCap systems, 3D scanners, LiDARs, and IMUs, offering increased accuracy and versatility (\Cref{3D_Dataset_APE}).

As discussed in \Cref{rgbapedataha}, researchers have traditionally relied on manual annotations or triangulation techniques to obtain 3D ground truth pose labels. An alternative approach for achieving high accuracy and low-noise 3D ground-truth animal pose data is to use a maker-based MoCap system with RGB cameras. For example, the Rat7M (2021) \citep{dunn2021geometric} and PFERD (2024) \citep{li2024poses} utilize MoCap system with RGB cameras and a set of makers to generate 3D ground truth pose data for rats and horses, respectively. Although these are effective for tracking individual animals from multiple views with temporal consistency, these datasets are limited by their focus on single animals and dependence on a large number of markers. To address these limitations, recent datasets like PAIR-R24M (2021) \citep{pairr2m} and 3D-POP (2023) \citep{Naik_2023_CVPR} have been developed. These datasets use MoCap systems with fewer markers (12 and 8 markers, respectively) to track the 3D postures of multiple animals in the same frame, offering more scalable solutions for multi-animal tracking. Similarly, a recent dataset \citep{RGBD_Dogs_2020_CVPR} utilizes MoCap systems with 63–82 markers to generate highly accurate 3D ground truth pose labels for dogs. This dataset also integrates depth sensors (\ie RGBD), which enable us to develop more flexible solutions for 3D APE without relying on physical markers. Although the MoCap-based dataset provides high-quality 3D ground truth pose labels, APE algorithms trained using these MoCap datasets may exhibit poor generalization to in-the-wild scenarios, as the data collection for the MoCap-based datasets is all done indoors in a controlled environment, as you can see in \Cref{3D_Dataset_APE}.

WildPose \citep{WildPose2024} is another promising multi-sensor system that combine RGB and LiDAR data. This system captures 2D RGB video and 3D point cloud data from a distance (up to 120 meters), making it particularly suited for studying free-ranging animals in natural habitats.

In addition to these technologies, infrared cameras for camera-trapping in the wild, which use infrared LEDs to illuminate the scene to enable nighttime observations, have also been employed for wild animal pose labelling. For instance, the LoTE-Animal (2023) \citep{LoTE_Animal_2023_ICCV} uses infrared trap cameras to collect large-scale video footage of endangered animals across 11 species. This dataset captures a wide range of variations—such as ecological seasons, weather conditions, viewpoints, and habitats—and provides both instance segmentation labels and 2D pose keypoint annotations. The LoTE-Animal dataset is especially valuable for advancing semi-supervised and self-supervised APE models, with potential applications in conservation efforts for endangered species.

Moreover, the VAREN (2024) dataset \citep{zuffi2024varen} focuses on 3D mesh representation, featuring 4k 3D scans of horses provided by the 3D scanner; this further advances 3D modelling and representation in APE research.

Lastly, the IMU-based animal pose dataset collected by \citep{yigit2022wearable} consists of acceleration and gyroscope measurements from sensors attached to the horse's limbs. The data captures each limb's 3D motion characteristics (like linear acceleration and angular velocity). It provides 3D information on movement and orientation but lacks absolute coordinates, as IMUs only capture relative motion.

These emerging sensor-based datasets offer new opportunities for APE, enhancing the precision and breadth of APE research across diverse animal species and environments.

\section{Error Metrics and Evaluation Methods} \label{EvaluationMethods}

This section provides insight into the error (err.) metrics and evaluation (eval.) process of the APE methods based on different sensors and modalities.

\subsection{RGB-based 2D Animal Pose Estimation Metric} \label{2D_RGB_metric}

This section discusses the common evaluation metrics used for RGB-based APE methods that extract poses in the 2D keypoint format, and more evaluation metrics are given in \Cref{2D_Image_APE_methods} and \Cref{2D_APE_Video_methods}.

2D single-animal pose estimation methods like DeepLabCut \citep{mathis2018deeplabcut} and LEAP \citep{pereira2019fast} use a common metric like Root Means Square Error (RMSE) to evaluate their performance. RMSE is versatile and can be used for both 2D and 3D pose estimation and for evaluating various types of keypoints, regardless of the specific application domain. In addition to RMSE, many APE methods also use the Probability of Correct Keypoint (PCK), which measures the fraction of joints where the predicted joint positions are within a certain threshold distance from the ground truth joint positions. It is a normalised measure that indicates how well the keypoints are localised, and a higher PCK value indicates better localisation performance. Single-animal pose estimation methods like Synthetic-to-Real \citep{li2021synthetic}, Kim \etal \citep{kim2022unified}, ScarceNet \citep{li2023scarcenet}, D-Gen \citep{li2023decompose} and FSKD \citep{lu2022few} use this metric for evaluation. Moreover, Li \etal \citep{li2020deformation} evaluate their single-animal APE methods with PCK, RMSE and Area Under Curve (AUC) along with an extension to these metrics with Permutation Invariant (PI). Li \etal \citep{li2020deformation} introduce PI metrics to evaluate the accuracy of pose estimation in challenging scenarios where it is difficult to uniquely identify limb identities (\eg in fruit flies) with traditional metrics. PI metrics are designed to account for the ambiguity in identifying corresponding limbs or joints between the predicted pose and ground truth. 

Object Keypoint Similarity (OKS) has also been used in evaluating 2D single-animal pose estimation methods like D-Gen \citep{li2023decompose}; 2D multiple-animal pose estimation methods like DepthFormer \citep{liu2022depthformer}, Straka \etal \citep{Straka_2024_WACV}, SLEAP \citep{pereira2022sleap}, and also 2D video based semi-supervised method like SemiMultiPose \citep{blau2022semimultipose}. OKS quantifies how well the keypoints are located to spatial scale and keypoint visibility, with closeness defined based on ground truth. The OKS lies in the range of 0 to 1, and a value close to 1 indicates that the detected keypoints coincide with the ground-truth keypoints. The OKS scores for every keypoint are computed for all objects in the dataset, and the average OKS is used to evaluate a model. Another common metric used for 2D APE is mean Average Precision (mAP), which has been used in evaluating ScarceNet \citep{li2023scarcenet}, SIPEC \citep{marks390deep}, SuperAnimal-Quadruped \citep{superanimal_2024}, OpenPose \citep{cao2017realtime}, DLCRNet \citep{lauer2022multi}, maDLC \citep{lauer2022multi}, SLEAP \citep{pereira2022sleap},  SuperAnimal-TopViewMouse \citep{superanimal_2024} and SemiMultiPose \citep{blau2022semimultipose}. The mAP is an evaluation metric based on OKS and has become widely used for pose estimation, particularly following its adoption in the COCO keypoint challenge. OKS measures the similarity between the predicted keypoints and the ground truth, and Average Precision (AP) is then calculated to assess the prediction's performance on a dataset by comparing it against different OKS thresholds. The mAP is the mean of several AP values, each calculated using a different threshold.

Last but not least, another metric that has been used for the 2D video-based APE method evaluation is Canonical Correlation Analysis (CCA), which is a statistical technique used to examine the relationship between two sets of variables. It determines linear combinations of the variables in each set that are maximally correlated; this metric has been used for evaluating the Lightning Pose \citep{biderman2023lightning} method, however, we have not seen wide adoption of this evaluation metric within the APE community.

\subsection{RGB-based 3D Animal Pose Estimation Metric}

This section discusses the common evaluation metrics used for both keypoint-based and mesh-based 3D RGB models within the APE community, more evaluation metrics for the 3D APE methods are given in \Cref{3D_Sing_Multi_view_APE_methods}.

3D APE metrics can also incorporate some of the metrics used in 2D APE. For example, PCK is employed in methods such as Three-D Safari \citep{zuffi2019three}, SMBLD \citep{biggs2020wldo}, Li \etal \citep{li2021coarse}, Birds of a Feather \citep{wang2021birds}, BITE \citep{ruegg2023bite}, MagicPony \citep{wu2023magicpony}, DigiDogs \citep{Digidogs_2024_WACV}, Monet \citep{yao2019monet}, DAN-NCE \citep{dunn2021geometric}, and Muramatsu \etal \citep{muramatsu2022improving}. The well-known RMSE metric is also used in 3D APE methods such as Hu \etal \citep{hu20233d}, Muramatsu \etal \citep{joska2021acinoset}, 3D-UPPER \citep{ebrahimi2023three}, and SBeA \citep{han2024multianimal}.

Apart from PCK and RMSE, 3D pose estimation-specific metrics such as Mean Per Joint Position Error (MPJPE) have been used in 3D APE methods like DigiDogs \citep{Digidogs_2024_WACV} and MAMMAL \citep{an2023three}. MPJPE measures the average Euclidean distance between the predicted and ground truth joint positions across all joints in the skeleton, quantifying the accuracy of the joint position estimation. A lower MPJPE indicates that the predicted joint positions are closer to the actual joint locations. An extension of the MPJPE metric is N-MPJPE, which has been used by Dai \etal \citep{dai2023unsupervised} for their 3D APE model. N-MPJPE involves normalising the ground truth 3D pose by subtracting the mean and dividing by the standard deviation, aligning the predicted 3D pose with the normalised ground truth using Procrustes alignment, and then calculating the MPJPE, which results in Normalized-MPJPE.

For mesh-based APE, a well-known evaluation metric is the 3D Chamfer Distance, which has been used in several model-free methods such as DOVE \citep{wu2023dove}, MagicPony \citep{wu2023magicpony}, and BANMo \citep{yang2022banmo}. The 3D Chamfer Distance measures the average distance between the reconstructed surface points and the ground truth surface points. It finds the nearest neighbour matches between the two sets of points and computes the distance, which is sensitive to outliers. A lower value indicates better reconstruction quality. 

Another approach in mesh-based APE is the model-based pose estimation methods, for example, SMAL \citep{zuffi20173d}. In these approaches, metrics like Scan-to-Mesh distance, keypoint reprojection error, and silhouette reprojection error are commonly used. The Scan-to-Mesh distance measures the accuracy of the reconstructed 3D shape by calculating the distance from each point on the scan to the closest point on the predicted mesh surface. Keypoint reprojection error evaluates the pose estimation by comparing the 2D projection of 3D keypoints from the predicted model with the annotated 2D keypoints in the images. It assesses how well the 3D pose model aligns with the 2D image data. Similarly, the silhouette reprojection error measures how accurately the 3D model silhouette matches the ground truth silhouette in the 2D image, indicating how well the 3D shape projection aligns with the visible boundaries of the animal in the image. Other model-based methods, such as SMBLD \citep{biggs2020wldo} and Li \etal \citep{li2021coarse}, use metrics like PCK and Intersection over Union (IoU) for performance evaluation.

Moreover, due to the scarcity of animal datasets that include 3D pose annotations, comprehensive quantitative evaluations of 3D pose predictions are not always feasible. Therefore, researchers often assess 3D pose predictions using qualitative means, as Sosa \etal \citep{sosa2023horse} explained. Examples of qualitative evaluations can be found in Salem \etal \citep{salem2019three}, Sosa \etal \citep{sosa2023horse}, Monet \citep{yao2019monet}, and Muramatsu \etal \citep{joska2021acinoset}.


\subsection{Other-Modality-based Animal Pose Estimation Metric}

Different sensor modalities have different ways of evaluating their APE models. For example, an IMU-based APE method \citep{yigit2022wearable} is evaluated by calculating the mean and standard deviation (Std) of the estimated joint angle errors and comparing the predictions with ground truth data obtained through a MoCap system. Additionally, the model is assessed for Intra-individual Variability (IAV) and Inter-individual Variability (IEV), which quantify the consistency of limb movement across different gait cycles and between different horses. Another metric used is the Variance Ratio (VR), which captures the repeatability of the gait cycle by evaluating how similar joint movements are within multiple cycles. This thorough approach ensures that this IMU-based model's performance is evaluated for accuracy and consistency across different conditions.

On the other hand, one of the RGBD-based APE methods, RGBD-Dog \citep{RGBD_Dogs_2020_CVPR}, is evaluated using MPJPE and PCK. The evaluation is conducted across different body parts and for entire dog poses. Additionally, Procrustes Analysis (PA) is applied to align the predicted skeleton with the ground truth skeleton. The PA is a statistical shape analysis technique. In the context of pose estimation, it aligns the predicted 3D skeleton with the ground truth skeleton by minimising the sum of squared distances between corresponding points in the two sets. In the RGBD-Dog \citep{RGBD_Dogs_2020_CVPR}, PA is applied to refine the existing PCK and MPJPE evaluation metrics to PA-PCK, PA-PCK 3D and PA-MPJPE, which is then used to measure the error after aligning the predicted and ground truth skeletons to focus purely on pose accuracy. Other multi-modal systems, for example, Sound of Motion (RGB+acoustic) \citep{li2022sound} and CLHOP (audio+video) \citep{CLHOP_Audio_video_2024}, also use PA-MPJPE for their APE evaluation in addition to PCK and IoU, which are common in the RGB-based APE evaluations.

This section briefly summarises some of the metrics for non-RGB-based APE models, more evaluation metrics are provided in \Cref{Multi_Sensor_APE}.

\section{RGB-based Animal Pose Estimation} \label{visionbased}

In recent years, APE has achieved significant progress. Among the various modalities utilised for APE, RGB-based approaches have emerged as the most developed and extensively studied research method \citep{pereira2019fast, cao2019cross, li2021synthetic, lauer2022multi}. These approaches leverage visual information captured by RGB cameras, take animal images or videos as inputs, and subsequently produce the animal's pose, which manifests as a 2D skeletal keypoints representation of pose, 3D skeletal keypoints representation of pose, or 3D animal mesh reconstructions, as exemplified in \Cref{fig:Skeleton Pose Representations} and \Cref{fig:Body Mesh Reconstructions}. In addition to skeletal keypoints and mesh representation for pose estimations, dense pose representations have also been used mainly for humans \citep{DensePose_2018_Humans} and chimpanzees \citep{DensePose_Chimpanzees}. In this paper, skeletal keypoints and mesh representations APE have been mainly analysed in detail, while keypoint-based methods remain dominant due to their established presence, easier access to annotated data, and lower computational requirements.

Before we go into the details of RGB-based APE methods, let us first discuss the development of this branch. Traditionally, we use direct visual observation to extract the animal poses, which can offer real-time and context-rich data. However, it suffers from inherent drawbacks, including observer bias, mental fatigue \citep{graving2019deepposekit}, and long data collection and observation times. Later, researchers start to use video monitoring and recording to expedite data collection and minimise the influence of human involvement \citep{graving2019deepposekit}. However, manual scoring of videos by eyes still remains time-consuming and often fails to capture subtle and fast-paced movements that are integral to various behaviours. Moreover, the reliance on human observers restricts the dataset size, this may also lead to statistical errors that limit researchers' ability to address scientific questions accurately.

Recently, automated body kinematics tracking techniques utilising advanced imaging software and hardware have been developed to tackle the challenges in manual scoring. Marker-based methods rely on attaching markers (\eg MoCap suits \citep{RGBD_Dogs_2020_CVPR}, animal-borne cameras \citep{patel2017tracking}) to animals for posture measurements. While these methods can achieve precise pose estimation, the reliance on physical markers can pose practical challenges, especially when investigating innate behaviours in natural habitats \citep{wu2023magicpony} or when observing the movements of small creatures like mice \citep{lauer2022multi} and fish. Furthermore, most marker-based approaches are limited to highly controlled lab environments and specialised hardware setups. Such methodologies are often impractical when the target animals demonstrate varied behaviours within unfamiliar environments.

With the rapid advancement in Computer Vision (CV), many automated markerless APE methods \citep{karashchuk2021anipose, mathis2018deeplabcut, lauer2022multi, pereira2019fast, graving2019deepposekit} start to employ supervised learning to directly infer and estimate animals' pose and movements from images or videos, removing the need for physical markers that may alter natural behaviour. For example, the markerless method DeepLabCut \citep{mathis2018deeplabcut} has been used for behaviour tracking in wild chimpanzees and bonobos \citep{wiltshire2023deepwild}.

However, a significant obstacle in APE is the acquisition of accurate ground truth poses as discussed in \Cref{datasetsection}. To address the scarcity of large-scale, well-labelled animal pose datasets, many weakly and semi-supervised \citep{cao2019cross}, semi-supervised \citep{kim2022unified, blau2022semimultipose, biderman2023lightning, yao2019monet}, self-supervised \citep{sosa2023horse, Digidogs_2024_WACV, yang2022banmo, han2024multianimal} and unsupervised learning \citep{dai2023unsupervised, kim2022unified, li2021synthetic, wu2023dove, ebrahimi2023three, wu2023magicpony, superanimal_2024, li2024learning} approaches have been developed. For instance, recent literature \citep{dai2023unsupervised} proposes an unsupervised 3D animal canonical pose estimation method with geometric self-supervision; this method necessitates only 2D poses for training, which significantly alleviates the requirement for extensively labelled data. Current APE research trends show growing interest in unsupervised, semi-supervised, and self-supervised methods due to the mentioned challenges to reach the same performance level as supervised methods. However, supervised methods are still a significant focus in current APE studies, especially with the emergence of larger datasets and benchmarking platforms like AP-10K \citep{yu2021ap}, APT-36K \citep{yang2022apt}, TigDog \citep{delpero15cvpr, delpero16ijcv}, and AcinoSet \citep{joska2021acinoset}, as we can see in \Cref{2D_Image_APE_methods}-\Cref{3D_Sing_Multi_view_APE_methods}.

This review paper categorises the existing uni-modal RGB-based APE studies into two main groups: 1) 2D APE methods from images (\Cref{2D_Image_APE_methods}, \Cref{2dframe}) and from videos (\Cref{2D_Image_APE_methods} \& \Cref{2D_APE_Video_methods}, \Cref{2D_APE_videos}), 2) 3D APE methods from images and videos (\Cref{3D_Sing_Multi_view_APE_methods}, \Cref{3dimagesandvides}).


\subsection{2D Animal Pose Estimation from an Image} \label{2dframe}

This section focuses on 2D APE methods that do not use temporal information to localize the 2D coordinates of an animal's or animal's joints from a single image frame (\Cref{2D_Image_APE_methods}). Note that the majority of the APE algorithms that can extract poses from videos (\ie multiple frames) can also extract poses from a single image. In \Cref{2D_Image_APE_methods}, column ``I,V" provides the information on APE methods that can be applied on both an image and videos; or only on an image.

Based on the number of target animals being processed in the scene, this pose estimation problem can be further divided into single-animal pose estimation (\Cref{singleanimal}) and multiple-animal pose estimation (\Cref{multianimal}). While pose estimation in individual animals has received significant attention and serves as the foundation for most APE studies, extending it to multiple-animal pose estimation is an active area of research that is still evolving \citep{pereira2022sleap}. 

\subsubsection{\textbf{Single Animal}} \label{singleanimal}

\begin{figure*}[!t]
\centering
\subfigure[Heatmap-based Method \citep{russello2022t}.]{\includegraphics[width=9cm, height = 4.6cm]{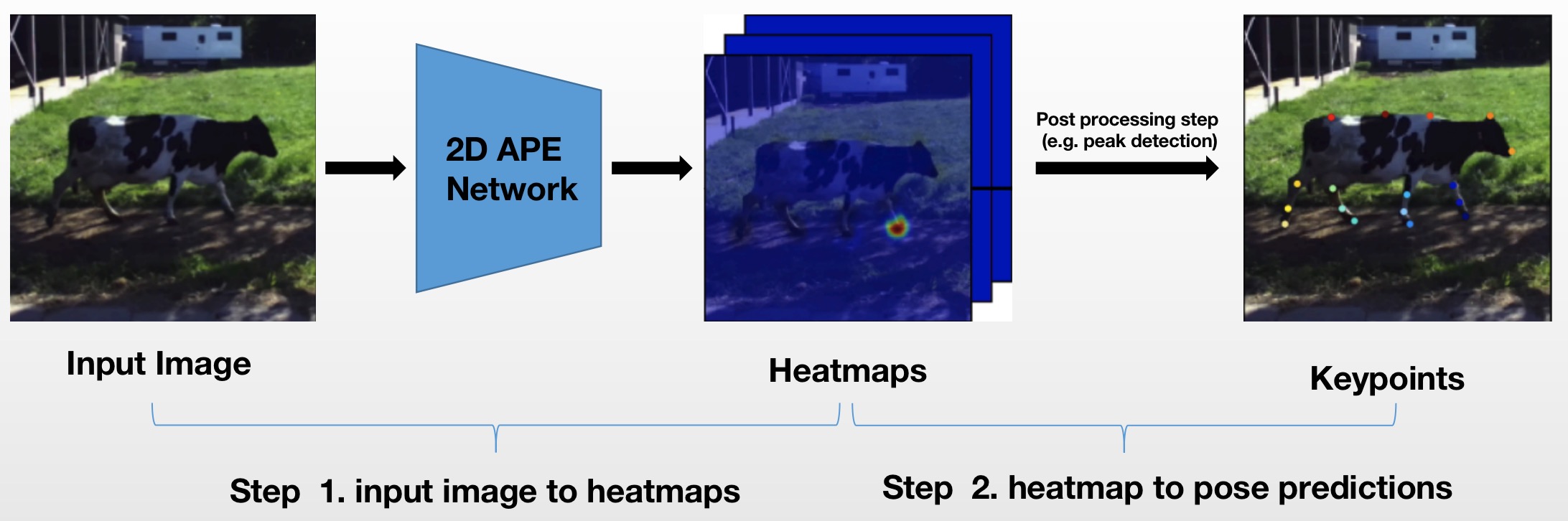}%
\label{heatheat}}
\hfil
\subfigure[Regression Method using Few-Shot Keypoint Detection \citep{lu2022few}.]{\includegraphics[width=8.2cm, height = 4.6cm]{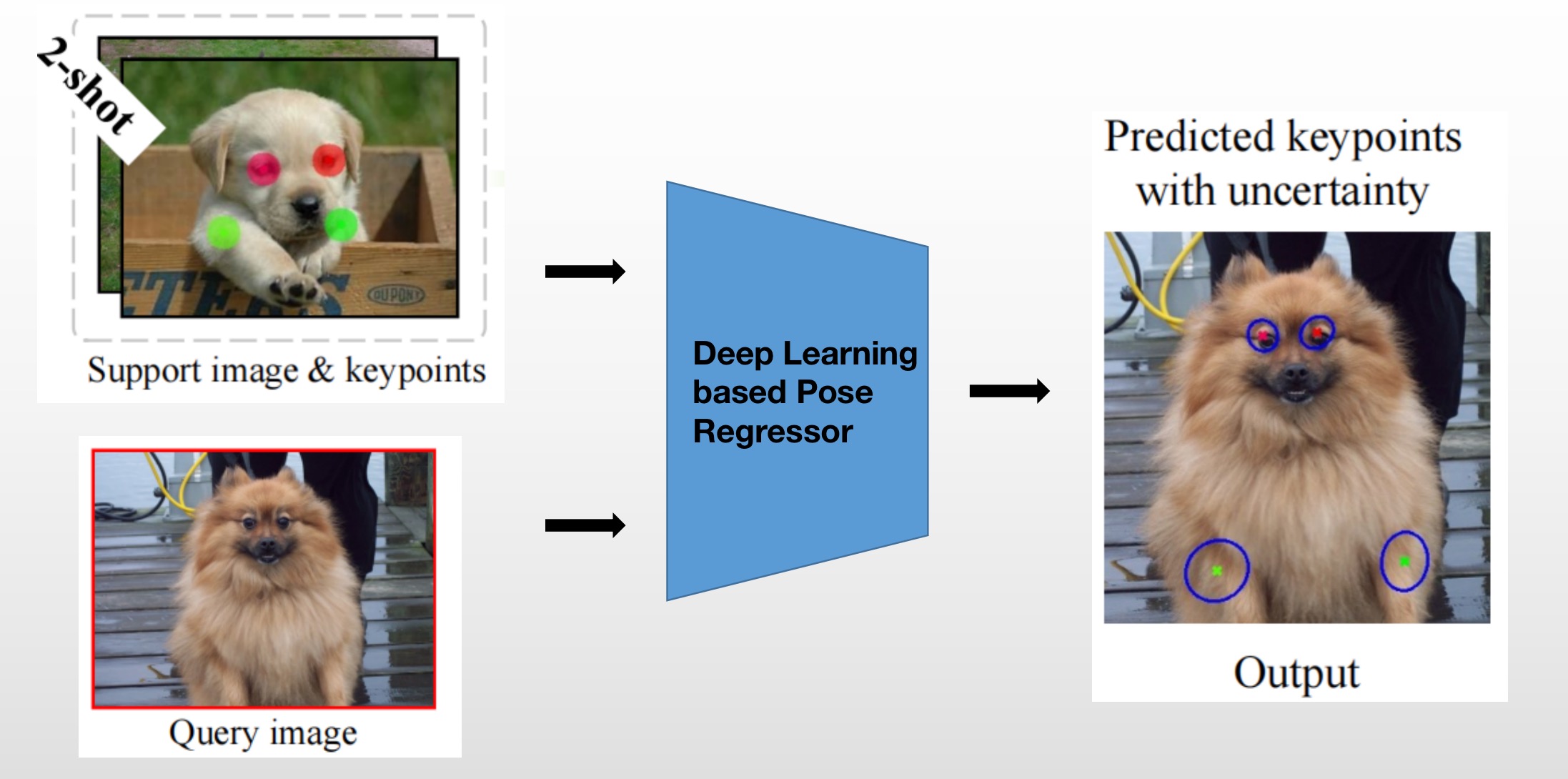}%
\label{RegressionRegression}}
\caption{The general architectures of 2D single-frame single-animal pose estimation methods. \centering}
\label{fig:single animal}
\end{figure*}

2D single-frame single-animal pose estimation aims to accurately estimate the 2D pose of an animal from a single image. These methodologies share a conceptual foundation with HPE \citep{zheng2020deep} and can be broadly classified into two categories: heatmap-based and regression methods. The former predicts the locations of an animal's body joints using heatmaps for supervision, necessitating a subsequent postprocessing step to identify the final keypoints based on the predicted heatmaps. In contrast, the latter approach directly regresses the keypoint coordinates of the animal's body joints, employing an end-to-end deep neural network to establish a mapping from the input image to the body keypoints. \Cref{fig:single animal} illustrates the general architecture of 2D single-frame single-animal pose estimation methods.

\textbf{Heatmap-based frameworks} such as \citep{pereira2019fast, kim2022unified, li2023scarcenet, li2020deformation, li2021synthetic, graving2019deepposekit, li2023decompose, mathis2018deeplabcut} have become popular in 2D APE. These approaches are structured around creating a 2D heatmap, interpreted as a probabilistic representation of body part locations within the image. The architecture of a typical heatmap-based scheme includes i) a CNN backbone to generate heatmaps from input images, followed by ii) a post-processing step to finalize the pose estimations based on these predictions (\Cref{heatheat}). There are some common architectures to input an image and output the corresponding heatmaps in step i); for instance, \citep{kim2022unified, li2021synthetic, li2023decompose, mathis2018deeplabcut} uses the ResNet \citep{he2016deep} and its variants to produce the heatmap; \citep{li2023scarcenet} adapts a variant of the HRNet \citep{sun2019deep} as the backbone; \citep{li2020deformation} applies the stacked hourglass \citep{newell2016stacked} architectures. Subsequently, in step ii), the predicted heatmaps are then processed to produce precise 2D spatial coordinates for each body joint using techniques such as peak detection \citep{graving2019deepposekit, mathis2018deeplabcut, pereira2019fast}. Although heatmap-based approaches are well known for their robustness against occlusions and variations in appearance, their computational complexity has been critiqued for impeding end-to-end training. Nonetheless, DeepPoseKit \citep{graving2019deepposekit} proposes an efficient end-to-end heatmap-based approach.

\textbf{Regression-based approaches} directly regress the coordinates of joint locations as continuous values, employing CNNs to learn a mapping from the input image to the joint coordinates. The output layer of the network consists of multiple units, each representing the $x$ and $y$ coordinates of a joint, and is trained to minimize the distance between the predicted and ground truth coordinates. Although regression-based approaches are less common than heatmap-based methods, they have shown great promise \citep{lu2022few}. One notable work \citep{lu2022few} adopts a regression-based approach to estimate keypoint positions directly, bypassing the intermediate heatmap representation (\Cref{RegressionRegression}). Their Few-shot Keypoint Detection (FSKD) model is designed to detect keypoints on unseen species with minimal annotated data, which is particularly useful in challenging scenarios where the species have not been previously encountered or the available annotated data is limited. Regression-based approaches offer the advantage of direct joint location estimation without additional peak detection steps, which can benefit real-time inference. However, they may face long-range 2D offset regression challenges, which may lead to  a notable performance gap compared to heatmap-based methods.

\begin{table*}[htbp]
\renewcommand{\arraystretch}{0.8}
\caption{RGB-based 2D APE Methods. In the column I,V: I,V is the methods that can perform pose estimation on both an image and videos, whereas, I is the APE methods that do only on an image. Column L indicates the learning types as follows: \textit{S} is supervised methods, \textit{U} is unsupervised methods, \textit{SF} is self-supervised methods, \textit{SM} is semi-supervised methods. $\symbifm$ is CNN architecture, $\symbmask$ is the use of the attention mechanism, $\symbgan$ is GAN, $\symbdb$ is underlying architecture as vision transformer, $\symbmultiscene$ is Multi-Scale Domain Adaptation Module (MDAM). $\symbposition$ is the pose estimation of the candidate algorithm given in the keypoint pose format.
$\symbbb $ is single-camera-view, $\symbvoxel$ is multi-camera-views. $\symbsurfels$ is indoor experimental set-up and $\symbdqb$  is in-the-wild experimental set-up. Symbols inside the \{\} indicate that both categories are used in that method. The method citation is available in column Year.}
\begin{center}
\small
\setlength{\tabcolsep}{1pt}
\begin{tabular}{c|c|c|c|c|c|c|c} 
\hline
\textbf{Sl} & \textbf{Year} & \textbf{Methods} & \textbf{I,V} & \textbf{L} & \textbf{Features} & \textbf{Evaluated On}& \textbf{Metrics}\\
\hline
\midrule
 \textcolor{blue}{} & \multicolumn{7}{c}{\textcolor{blue}{2D Single-Animal Pose Estimation Methods}} \\
 \midrule
 & & \textbf{Heatmap-based:}\\
 1 & \citeyear{mathis2018deeplabcut} & DeepLabCut  & I,V & \textit{S} & $\symbposition$, $\symbifm$, $\symbbb$, $\symbsurfels$ & Mammals & RMSE (pixels)\\
  
 2 & \citeyear{pereira2019fast} & LEAP  & I,V & \textit{S} &  $\symbposition$, $\symbifm$, $\symbbb$, $\symbsurfels$ & Fly, Mouse & RMSE (pixels)\\
 
 3 & \citeyear{graving2019deepposekit} & DeepPoseKit  & I,V & \textit{S} & $\symbposition$, $\symbifm$, $\symbbb$, \{$\symbsurfels$, $\symbdqb$\}  & Zebra (\citeyear{zuffi2019three}) ($\symbdqb$); &  Euclidean Error,  \\
 & & & & & & Desert locust, Fly ($\symbsurfels$) & Bayesian Inference \citeyear{graving2019deepposekit}\\
 
 4 & \citeyear{li2020deformation} & Li \etal  & I &  \textit{S} & $\symbposition$, $\symbifm$,  $\symbbb$, $\symbsurfels$ & Fly, Worm, Zebrafish & RMSE, PI-RMSE, PCK, AUC, \\
  & & & & & &  & PI-PCK, PI-AUC\\
  
 5 & \citeyear{li2021synthetic} & Synthetic-to-Real  & I & \textit{U} & $\symbposition$, $\symbmultiscene$,  $\symbbb$, $\symbdqb$ & Horse, TigDog (\citeyear{delpero16ijcv}) & PCK@0.05 (\%) \\
 
 6 & \citeyear{kim2022unified} & Kim \etal  & I &  \textit{SM,U} & $\symbposition$, $\symbifm$,  $\symbbb$ \{$\symbsurfels$, $\symbdqb$\} & TigDog (\citeyear{delpero16ijcv}),  & PCK@0.05 (\%)\\
 & & & & & & Animal Pose (\citeyear{cao2019cross}),  \\
 & & & & & & Synthetic Animal (\citeyear{Mu_2020_CVPR})  \\

 7 & \citeyear{li2023scarcenet} & ScarceNet  & I & \textit{S} & $\symbposition$, $\symbifm$,  $\symbbb$, $\symbdqb$ &  TigDog (\citeyear{delpero16ijcv}), & mAP (\%), PCK@0.05 (\%) \\
  & & & & & & AP-10K (\citeyear {yu2021ap})  \\

 8 & \citeyear{li2023decompose} & D-Gen  & I &  \textit{S} & $\symbposition$, $\symbmask$, $\symbbb$, $\symbdqb$ & Animal Pose (\citeyear{cao2019cross}), & OKS, PCK@0.05 (\%)\\
 & & & & & & AP-10K (\citeyear {yu2021ap}), \\
 & & & & & & Animal Kingdom (\citeyear{ng2022animal})\\
 
\hline
 & & \textbf{Regression-based:}\\ 
 
 9 & \citeyear{lu2022few} & FSKD  & I & \textit{S} & $\symbposition$, $\symbifm$, $\symbbb$, \{$\symbsurfels$, $\symbdqb$\} &  Animal Pose (\citeyear{cao2019cross}) &  PCK\\

\midrule
 \textcolor{blue}{} & \multicolumn{7}{c}{\textcolor{blue}{2D Multi-Animal Pose Estimation Methods}}\\
 \midrule
 & & \textbf{Top-down:} \\
 

   10 & \citeyear{marks390deep} & SIPEC  & I,V & \textit{S} & $\symbposition$, $\symbifm$,  $\symbbb$, $\symbsurfels$ & Primates, Mice & Pearson Correlation, RMSE (pixels), \\
 & & & & & &&  mAP (\%), IoU, Dice Coefficient\\ 
 
 11 & \citeyear{liu2022depthformer} & DepthFormer  & I & \textit{S} & $\symbposition$, $\symbdb$, $\symbbb$, $\symbdqb$ & AP-10K (\citeyear{yu2021ap}) & OKS\\

 12 & \citeyear{Straka_2024_WACV} & Straka \etal  & I & \textit{S} & $\symbposition$, $\symbifm$, $\symbbb$, $\symbdqb$ & Lynx & PCK, OKS, AP \\

 13 & \citeyear{superanimal_2024} & SuperAnimal-  & I,V & \textit{S} & $\symbposition$, $\symbifm$, $\symbbb$, $\symbdqb$ & Animal Pose (\citeyear{cao2019cross}), & RMSE (pixels), Normalized Error, \\
 & & Quadruped  & & & &  AP-10K (\citeyear{yu2021ap}) & mAP  \\
 \hline

 & & \textbf{Bottom-up:} \\
 
 14 & \citeyear{cao2017realtime} & OpenPose  & I,V  & \textit{S} & $\symbposition$, $\symbifm$, $\symbbb$, \{$\symbsurfels$, $\symbdqb$\} & COCO (\citeyear{lin2014microsoft}), MPII (\citeyear{andriluka20142d}) & mAP (\%)\\

 15 & \citeyear{blanco2021multiple} & Negrete \etal  & I & \textit{S} & $\symbposition$, $\symbifm$, $\symbbb$, $\symbdqb$ & MSCOCO (\citeyear{lin2014microsoft}), & AP \\
  & &  & & & &  MacaquePose (\citeyear{MacaquePose_2021}) &   \\

 16 & \citeyear{farahnakian2021multi} & Farahnakian \etal  & I,V & \textit{S} & $\symbposition$, $\symbifm$, $\symbbb$, $\symbdqb$ & Pig (\citeyear{Nath2019}) & Error (pixels), RMSE (pixels) \\

  17 & \citeyear{lauer2022multi} & DLCRNet, maDLC  & I,V & \textit{S} & $\symbposition$, $\symbifm$,  $\symbbb$, \{$\symbsurfels$, $\symbdqb$\} & 3 Mice, 2 Marmosets ($\symbdqb$); & RMSE (pixels), mAP, PCK \\
   & &  & & & &  1 Mice ($\symbsurfels$)  &   \\
 \hline
 & & \textbf{Bottom-up \& } \\
  & & \textbf{Top-down:} \\


  18 & \citeyear{pereira2022sleap} & SLEAP  & I,V & \textit{S} & $\symbposition$, $\symbifm$, $\symbbb$, $\symbsurfels$ &  Flies, Bees, Mice, Gerbils &  mAP (\%), PCK, OKS, Localization \\

  19 & \citeyear{superanimal_2024} & SuperAnimal-  &  I,V & S & $\symbposition$, $\symbifm$, $\symbbb$, $\symbdqb$ & Animal Pose (\citeyear{cao2019cross}), & RMSE (pixels), Normalized Error, \\
     & & TopViewMouse & & & &  AP-10K (\citeyear {yu2021ap}) & mAP  \\

\hline
\hline
\end{tabular}
\label{2D_Image_APE_methods}
\end{center}
\end{table*}

\subsubsection{\textbf{Multiple Animals}} \label{multianimal}

\begin{figure*}[!t]
\centering
\subfigure[Top-Down Approach: SLEAP Method \citep{pereira2022sleap} Instances - Parts.]{\includegraphics[width=9cm, height = 3cm]{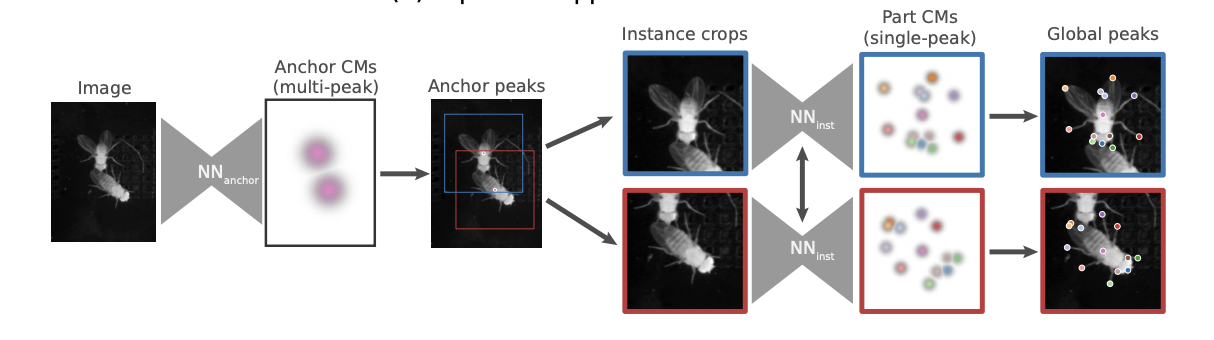}%
\label{aaa}}
\hfil
\subfigure[Bottom-Up Approach: SLEAP Method \citep{pereira2022sleap} Parts - Instances.]{\includegraphics[width=9cm, height = 3cm]{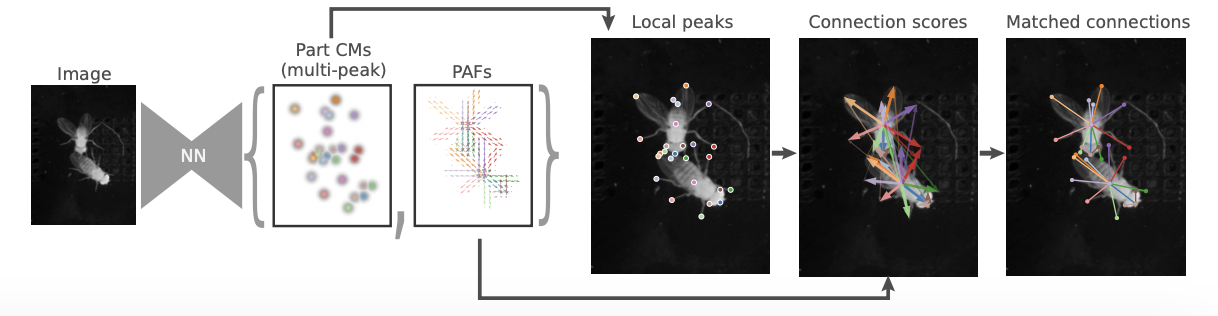}%
\label{bbb}}
\caption{2D single-frame multiple-animal pose estimation using SLEAP method \citep{pereira2022sleap}. \centering}
\label{fig:multi animal}
\end{figure*}

We have seen significant developments in markless pose estimation in individual animals. However, extending these methodologies to multiple animals poses unique challenges, particularly in studying social behaviours or animals in their natural environments \citep{yang2022apt, gong2022multicow}. For example, it is common for the body parts of one animal to be obscured by another in a multi-animal pose estimation. Another challenge arises from scale variation, where animals at different distances from the camera show differences in apparent sizes. Inter-species variation also introduces additional complexity due to differences in body structures and poses. However, given these challenges, 2D multiple-animal pose estimation research has still garnered significant attention due to its potential applications in various domains, such as animal behaviour analysis, wildlife monitoring, and ecological studies. Methodologies developed in this field are categorized into two main categories: top-down methods \citep{marks390deep, liu2022depthformer, Straka_2024_WACV, superanimal_2024} and bottom-up methods \citep{cao2017realtime, blanco2021multiple, farahnakian2021multi, lauer2022multi} (\Cref{fig:multi animal}).

\textbf{Top-down approaches} for multiple-animal pose estimation (\Cref{aaa}) generally involve a two-step process: first i) detecting each animal in the input image using pre-trained object detectors and then ii) applying 2D single-animal pose estimation techniques to estimate the pose for each detected instance. Common object detectors in step i) include Faster R-CNN \citep{ren2015faster}, the Single Shot MultiBox Detector (SSD) \citep{liu2016ssd}, You Only Look Once (YOLO) v4 \citep{gong2022multicow}, and RetinaNet \citep{lin2017focal}; common 2D APE methods in step ii) includes variants of ResNet, HRNets and transformer. In the early work, SIPEC \citep{marks390deep} uses SIPEC:SegNet (a Mask R-CNN architecture \citep{he2017mask}) to segment animal instances in the images or videos first, then it uses SIPEC:PoseNet to estimate the pose for each individual. Later, we can spot a trend of using HRNets and transformers as the 2D APE backbone architecture in step ii) to achieve better performance, especially with the limited animal pose data. For example, the top-down DepthFormer method \citep{liu2022depthformer} has been proposed to achieve lightweight multiple-animal pose estimation by designing a new basic block based on the transformer structure and the similarities between self-attention and depthwise convolution. In 2024, SuperAnimal-TopViewMouse \citep{superanimal_2024} adapts Faster R-CNN for enhanced feature extraction to detect animal instances in the given image or videos first, and then also uses the transformer-based AnimalTokenPose models to achieve high zero-shot APE performance in OOD animal datasets. In the same paper, both SuperAnimal-TopViewMouse \citep{superanimal_2024} and SuperAnimal-Quadruped \citep{superanimal_2024} models also enable us to use HRNet-w32 \citep{wang2020deep} as the underlying architecture to get the final animal 2D pose. Until now, SuperAnimal \citep{superanimal_2024} has claimed to be the current state-of-the-art across a wide variety of animal species for both 2D images and videos. More information on how SuperAnimal is adopted to videos is discussed in \Cref{2D_APE_videos}. Another recent top-down approach for 2D endangered species (i.e. lynx) pose estimation with limited data \citep{Straka_2024_WACV} also leverages the high-resolution network HRNet-w32 \citep{wang2020deep} architecture in step ii) to get accurate 2D animal pose, as this HRNet-w32 architecture helps to maintain detailed spatial information throughout the network. All of these methodologies prove the trend.

In contrast, \textbf{bottom-up methods} (\Cref{bbb}) prioritize predicting all keypoint locations in an image first, then grouping them into individual instances. Initial work in this area includes methods like OpenPose \citep{cao2017realtime}, which introduced the concept of Part Affinity Fields (PAFs) to associate body parts with individuals coherently. This PAFs idea was further adapted in the DLCRNet and maDLC architecture \citep{lauer2022multi} to predict the location and orientation of animal limbs and link the specific keypoints to the correct individual. OpenPose \citep{cao2017realtime} was also used in \citep{blanco2021multiple} to associate detected keypoints to particular individuals by taking advantage of similarities in skeleton structure between humans and monkeys. Further refinement in the bottom-up approach was made by works such as \citep{farahnakian2021multi}, which focused on pig body part association using DeepLabCut \citep{mathis2018deeplabcut}. 

Both top-down and bottom-up methods have limitations and features. For example, top-down approaches are generally less sensitive to scale variance of objects compared to bottom-up methods, as they use bounding boxes to identify each individual first, then estimate the joints of the detected body in that box region. However, they often face difficulties in scenarios involving occlusions, as the detection step may fail to identify all instances accurately. On the other hand, the primary shortcoming of the bottom-up approaches often stems from incorrect assembly. To allow the users to compare the two competing approaches directly, researchers have developed both bottom-up and top-down variants for maDLC \citep{lauer2022multi}, SuperAnimal-TopViewMouse \citep{superanimal_2024} and SLEAP \citep{pereira2022sleap}. The decision between these two methods depends on the specific requirements and constraints of the application.

Before we move on to the next section, it is worth noting that some of the APE methods explained in \Cref{2dframe} can perform pose estimation on both images and videos, as can be seen in \Cref{2D_Image_APE_methods} in column IV. In \Cref{2D_APE_videos}, we explain which of these video-based 2D APE methods extract pose frame by frame, and which uses the spatiotemporal signals of video data in an end-to-end manner for pose estimation.


\subsection{2D Animal Pose Estimation from Videos} \label{2D_APE_videos}

2D multi-frame APE transforms consecutive image sequences or videos into probabilistic estimates of the 2D spatial coordinates of keypoint locations for target animals across frames. Based on how these approaches leverage the temporal dimension in video data to extract poses, we categorize 2D APE from videos into two main categories: (1) post-processing methods to incorporate temporal information; and (2) end-to-end methods to incorporate temporal information, as shown in \Cref{2D_APE_Video_methods}.


\begin{table*}[htbp]
\renewcommand{\arraystretch}{1}
\caption{RGB-based 2D APE methods from videos. All the symbols and column meanings are defined in \Cref{2D_Image_APE_methods}, and any additional symbols are defined here: $\symbtensorfac$ is the underlying architecture as 3D-CNN, $\symbifm$+$\symbvideo$ is the use of CNN-based pose module followed by an Optical Flow module, $\symbnf$ is Bayesian Ensembling, $\symbdepth$ is Probabilistic Graphical Model. The method citation is available in column Year.}
\begin{center}
\small
\setlength{\tabcolsep}{1pt}
\begin{tabular}{c|c|l|c|c|c|c} 
\hline
\textbf{Sl} & \textbf{Year} & \textbf{Methods} & \textbf{L} & \textbf{Features} & \textbf{Evaluated On}& \textbf{Metrics}\\
\hline




    \midrule
    \textcolor{blue}{} & \multicolumn{6}{c}{\textcolor{blue}{2D APE from Videos:  End-to-End Methods to Incorporate Temporal Information}}\\
    \midrule
    
    1 & \citeyear{liu2021optiflex} & OptiFlex   & \textit{S} & $\symbposition$, \{$\symbifm$+$\symbvideo$\}, \{$\symbbb$, $\symbvoxel$\}, $\symbsurfels$ & Mouse, Fish, Monkey, Fly & aPCK error (\%)\\

    2 & \citeyear{russello2022t} & T-LEAP  & \textit{S} & $\symbposition$, $\symbtensorfac$, $\symbbb$, $\symbdqb$ & Cow & PCKh@0.2(\%)\\

    3 & \citeyear{pereira2022sleap} & SLEAP  & \textit{S} & $\symbposition$, $\symbifm$, $\symbbb$, $\symbsurfels$ &  Flies, Bees, Mice, Gerbils &  mAP (\%), PCK, OKS, Localization \\

    4 & \citeyear{blau2022semimultipose} & SemiMultiPose  & \textit{SM} & $\symbposition$,  \{$\symbifm$+$\symbdepth$\},
    $\symbvoxel$, $\symbsurfels$ & Mouse, HoneyBee, Fly & mAP, OKS\\

    
    5 & \citeyear{biderman2023lightning} & Lightning Pose  & \textit{SM} & $\symbposition$, $\symbnf$, $\symbvoxel$, $\symbsurfels$ & Mouse, Fish & CCA, Pixel Error, AUC\\

    6 & \citeyear{superanimal_2024} & SuperAnimal  & \textit{U} & $\symbposition$, $\symbifm$, $\symbbb$, $\symbdqb$ & Animal Pose (\citeyear{cao2019cross}), & RMSE (pixels), Normalized Error, \\
    &  &  &  &  &  AP-10K (\citeyear {yu2021ap}) & mAP \\

\hline
\hline
\end{tabular}
\label{2D_APE_Video_methods}
\end{center}
\end{table*}

\subsubsection{\textbf{Post-Processing Methods to Incorporate Temporal Information}} \label{fbfAPE}

Early APE frameworks process each video frame independently to predict pose keypoints, without inherently incorporating temporal features in an end-to-end manner during training. Temporal consistency is typically addressed through optional post-processing methods, such as smoothing or filtering, to improve performance. Some examples that follow this frame-by-frame approach include DeepLabCut \citep{mathis2018deeplabcut}, DeepPoseKit \citep{graving2019deepposekit}, and LEAP \citep{pereira2019fast}.

\subsubsection{\textbf{End-to-End Methods to Incorporate Temporal Information}}

Recently, researchers have achieved more robust predictions by utilising the spatiotemporal information embedded within video data during pose estimation model training in an end-to-end manner. By evaluating inherent temporal coherence and ensuring consistency across successive frames during training, models can produce more accurate and stable pose estimations over time. We highlight several methods that fall in this category in the following bullet points and \Cref{2D_APE_Video_methods}.

\textbf{Extensions of LEAP Method:} A notable extension of the LEAP \citep{pereira2019fast} method is T-LEAP \citep{russello2022t}, which conceptualises temporal information as an auxiliary dimension within the CNN framework, thereby employing 3D convolutional layers to capture both spatial and temporal features concurrently to extract animal poses from video data. To leverage the temporal information embedded in videos, T-LEAP processes a sequence of frames of length $T$, and outputs the confidence maps linked to the last frame $t=T$. As for architectural modifications compared to a deeper version of LEAP, all convolution, pooling and transposed-convolution operations transition from 2D to 3D. Notably, these 3D convolutional operations are extended to not only the spatial dimension of the input but also to the temporal dimension, thus facilitating the extraction of spatiotemporal signals from the video data, and making T-LEAP robust for occlusions. For example, in a scenario where a cow's leg is discernible in one frame but obscured in the subsequent frame, T-LEAP can employ visible information to predict the probable position of the leg in the occluded frame. Another extension of the LEAP model is SLEAP \citep{pereira2022sleap}, which also incorporates temporal information, but unlike LEAP, this extends the model to multi-animal pose estimation.

\textbf{Optical Flow-based Methods:} OptiFlex \citep{liu2021optiflex} is an optical flow-based methods that allow the estimation of motion vectors between consecutive frames to capture the temporal dynamics of animal poses. In particular, OptiFlex \citep{liu2021optiflex} combines a deep learning-based pose estimation module called FlexibleBaseline with an optical flow module named OpticalFlow to enhance its APE performance across multiple frames. The FlexibleBaseline module predicts the initial pose for each frame, and the OpticalFlow module converges initial predictions on a target frame and its adjacent frames into a final prediction for the target frame. By tracking the pixel-level displacements over time, optical flow estimation techniques, such as the Lucas-Kanade algorithms \citep{baker2004lucas}, can compute dense motion fields between frames, thereby inferring the movement of body parts and joints. Suppose a certain body part is occluded or unclear in a frame. In that case, the model can infer its likely position based on the detected motion and poses in adjacent frames, showing its robustness against temporary obstruction of keypoint poses.

\textbf{Future Trend using Semi-supervised and Unsupervised Video Adaptation Learning:}
Recently, there has been a growing shift towards using semi-supervised learning (\eg SemiMultiPose \citep{blau2022semimultipose}, Lightning Pose \citep{biderman2023lightning}) and unsupervised video adaptation learning (\eg SuperAnimal \citep{superanimal_2024}) to extract poses from videos. In 2D single-frame APE methods, supervised learning dominates, relying on many labelled frames for training. However, manually adding annotations for videos is costly and labour-intensive, especially for multiple instances. In real-world scenarios like wildlife videography, there are exponentially more abundant unlabeled frames than labelled frames. Given these deficiencies, SemiMultiPose \citep{blau2022semimultipose} proposes a semi-supervised architecture to exploit the abundant structures pervasive in these unlabeled video frames to enhance the training process, which is particularly useful in the sparse labelling video-based scenarios. Similarly, Lightning Pose (2023) \citep{biderman2023lightning} incorporates a semi-supervised approach to exploit the spatiotemporal statistics present in unlabeled videos. This methodology introduces unsupervised training objectives that penalise the network when its predictions deviate from the expected smoothness of physical motion, multiple-camera-views geometry, or stray from a low-dimensional subspace of plausible body configurations. Additionally, Lightning Pose predicts the pose for a specific frame by considering the temporal context from surrounding unlabeled frames. This incorporation of context is instrumental in resolving transient occlusions and ambiguities between proximal and morphologically similar body parts. Last, the SuperAnimal (2024) \citep{superanimal_2024}, an unsupervised video adaptation APE method, can be applied to images and videos. However, its approach differs depending on the data type. For images, SuperAnimal operates as a supervised method, relying on labelled data for training. In contrast, for videos, it functions as an unsupervised method by incorporating unsupervised techniques, such as video adaptation and pseudo-labelling, to improve performance on unseen videos. This method helps reduce temporal jitter and refine model outputs without further manual annotation. However, the unsupervised techniques are applied after the model has been initially trained in a supervised manner. These works demonstrate better APE performance with fewer labels, generalise better to unseen videos, and provide smoother and more reliable pose trajectories for downstream analysis, which proves the shift towards semi-supervised and unsupervised learning in APE.

\subsection{3D Animal Pose Estimation from Images and Videos} \label{3dimagesandvides}

3D APE refers to the task of determining the locations of animal body joints in 3D space. This challenge extends beyond the 2D APE by adding depth information to the analysis, which allows for a deeper understanding of animal behaviour, biomechanics, and motion. Some 3D APE methods extract poses from monocular imagery and video data streams. However, these singular perspectives introduce an ill-posed nature to the problem, as the reduction from 3D reality to 2D representation inherently results in the loss of spatial data. Conversely, the integration of multiple-camera-views imaging facilitates a well-conditioned approach to interpolate the lost dimension for 3D APE using geometric consistency. Based on the number of camera views used to extract the animal poses, we categorize the 3D APE into two sections: (1) 3D APE from single-camera-view images and videos \citep{salem2019three, gosztolai2021liftpose3d, dai2023unsupervised, sosa2023horse, Digidogs_2024_WACV, hu20233d, zuffi20173d, zuffi2019three, biggs2020wldo, li2021coarse, wang2021birds, ruegg2022barc, wu2023dove, ruegg2023bite, wu2023magicpony, yang2022banmo, li2024learning}, (2) 3D APE from multiple-camera-views images and videos \citep{yao2019monet, bala2020automated, zhang2021animal,karashchuk2021anipose,  dunn2021geometric, muramatsu2022improving, ebrahimi2023three, an2023three, han2024multianimal}.


\begin{table*}[ht]
\renewcommand{\arraystretch}{0.8}
    \caption{3D RGB-based APE methods. All the symbols and column meanings are defined in \Cref{2D_Image_APE_methods} and \Cref{2D_APE_Video_methods}, and any additional symbols are defined here: $\symbei$ is shape model \citep{zuffi20173d}, $\symbcm$ is Structured Forests architecture \citep{Structured_Forests_2015}, $\symboctree$ is Graph Convolutional Network (GCN), $\symbpc$ is Stacked Hourglass architecture based on CNN, $\symbgan$ is GAN, $\symbedg$ is the combination of Neural Radiance Fields (NeRF), Volumetric Signed Distance Functions (VolSDF) \& Differentiable Physics Simulation, $\symbshape$ is a framework based on Epipolar Divergence, $\symblidar$ is Gaussian-based model,  $\symbsemanticsegmentation$ is Statistical Shape Model (SSM), $\symbevent$ is NeRF, $\symbscrew$ is Semantic Bank of Skinned Models (SBSM), $\symbfd$ is feed-forward neural network. $\symbmesh$ is the Body Mesh Recovery method. The method citation is available in column Year.}
\begin{center}
\small
\setlength{\tabcolsep}{1pt}
\begin{tabular}{c|c|l|c|c|c|c|c} 
\hline
\textbf{Sl} & \textbf{Year} & \textbf{Methods} & \textbf{VI} & \textbf{L} & \textbf{Features} & \textbf{Evaluated On} & \textbf{Evaluation Metrics}\\
\midrule
\textcolor{blue}{} & \multicolumn{7}{c}{\textcolor{blue}{ 3D APE from Image \& Video - Single-Camera-View Methods}} \\
 \midrule
  & & \textbf{Skeleton-Only:}\\

   1 & \citeyear{salem2019three} & Salem \etal  & I,V & \textit{S} & $\symbposition$, $\symbcm$, $\symbbb$, $\symbsurfels$ & Mouse & Pose
  Distance Measure (\citeyear{cascaded_pose_regression_2020}), \\
 & & & & & & & Reconstruction Err.  \\

  2 & \citeyear{gosztolai2021liftpose3d} & LiftPose3D  & V & \textit{S} & $\symbposition$, $\symbifm$, $\symbbb$, $\symbsurfels$ & Mice, Rats, Macaques & MAE \\

  3 & \citeyear{hu20233d} & Hu \etal  & V & \textit{S} & \{$\symbposition$, $\symbmesh$\}, $\symbpc$, $\symbbb$, $\symbsurfels$ & Mouse & OKS for 2D Keypoints, \\
   & & & & & & & RMSE for 3D Keypoints  \\

 4 & \citeyear{dai2023unsupervised} & Dai \etal  & I & \textit{U} & $\symbposition$, $\symbgan$, $\symbbb$, $\symbdqb$  & SMAL (\citeyear{zuffi20173d}),  & N-MPJPE \\
  & & &  & &  &  AcinoSet (\citeyear{joska2021acinoset})  &  \\
 
 5 & \citeyear{sosa2023horse} & Sosa \etal  & I & \textit{SF} & $\symbposition$, $\symbifm$, $\symbbb$, $\symbdqb$ & Horse - Weizmann (\citeyear{Weizmann_dataset_2004}) & PCK@0.05 for 2D,\\
   & & & & & & & Qualitative for 3D  \\

 6 & \citeyear{Digidogs_2024_WACV} & DigiDogs/D-Pose  & I & \textit{SF} & $\symbposition$, $\symbdb$, $\symbbb$, $\symbdqb$ & Dogs & PCK for 2D \& 3D, \\
    & & & & & & & MPJPE for 3D \\

    \hline
  & & \textbf{AMR}\\
  & & \textbf{(Model-Based):}\\

  7 & \citeyear{zuffi20173d} & SMAL  & I & \textit{S} & $\symbmesh$, $\symbei$, $\symbbb$, \{$\symbsurfels$, $\symbdqb$\} & Quadruped & Scan-to-Mesh Distance, \\
  & & & & & && Keypoint and Silhouette \\
  & & & & & && Reprojection Err. \\

  8 & \citeyear{zuffi2019three} & Three-D Safari  & I & \textit{S} & $\symbmesh$, $\symbei$, $\symbbb$, $\symbdqb$ & Zebra &  PCK, IoU, Geodesic Distance\\

  9 & \citeyear{biggs2020wldo} &  SMBLD  & I & \textit{S} & $\symbmesh$,  $\symbei$, $\symbbb$, \{$\symbsurfels$, $\symbdqb$\} & StanfordExtra (\citeyear{biggs2020wldo}) & IoU, PCK \\

 10 & \citeyear{li2021coarse} & Li \etal  & I & \textit{S} & $\symbmesh$, \{$\symbei$+$\symboctree$\} $\symbbb$, \{$\symbsurfels$, $\symbdqb$\} & BADJA (\citeyear{BADJA_biggs2018creatures}), & PCK, IoU\\
   & & &  & &  &  Animal Pose (\citeyear{cao2019cross}),   &  \\

 & & & & & & StanfordExtra (\citeyear{biggs2020wldo})\\

  11 & \citeyear{wang2021birds} & Birds of a Feather   & I & \textit{S} & $\symbmesh$, $\symbei$, $\symbbb$, $\symbdqb$ & Birds, & PCK, IoU\\
   & & & & & & StanfordExtra (\citeyear{biggs2020wldo})\\

  12 & \citeyear{ruegg2022barc} & BARC  & I & \textit{S} & $\symbmesh$, \{$\symbei$+$\symbpc$\}, $\symbbb$, \{$\symbsurfels$, $\symbdqb$\} & StanfordExtra (\citeyear{biggs2020wldo}) & PCK \& IoU for 2D, \\
  &&&&&&& Breed Prototype Consistency \&  \\
  &&&&&&& Perceptual Shape Eval. for 3D  \\

  13 & \citeyear{ruegg2023bite} & BITE  & I & \textit{S} & $\symbmesh$, $\symbei$, $\symbbb$, \{$\symbsurfels$, $\symbdqb$\} &  StanfordExtra (\citeyear{biggs2020wldo}), & Scan-to-Mesh Distance,\\ 
  &&&&&& Semi-Synthetic 3D (\citeyear{ruegg2023bite}) & Mesh-to-Scan Distance, \\
     & & & & & & & PCK, IoU\\

 \hline
 & & \textbf{AMR}\\
  & & \textbf{(Model-Free):}\\

 14 & \citeyear{yang2022banmo} & BANMo  & I,V & \textit{SF} & $\symbmesh$, \{$\symbei$+$\symbevent$\} , $\symbvoxel$, \{$\symbsurfels$, $\symbdqb$\} & Animals \& Humans & F-score,\\
     & & & & & & & 3D Chamfer Distance (cm)\\

 15 & \citeyear{wu2023dove} & DOVE  & I,V & \textit{U} & $\symbmesh$, $\symbifm$, $\symbbb$, $\symbdqb$ & Bird, Horse,  & mIoU, \\
    & & & & & & 3D Toy Bird Data & 3D Chamfer Distance (cm)\\

 16 & \citeyear{wu2023magicpony} & MagicPony  & I & \textit{U} & $\symbmesh$, $\symbdb$, $\symbbb$,  $\symbdqb$ &  COCO (\citeyear{lin2014microsoft}), & PCK, \\
     & & & & & & Horse-10 (\citeyear{mathis2021pretraining}), & 3D Chamfer Distance (cm) \\
      &&&&&& DOVE (\citeyear{wu2023dove}), others \\


  17 & \citeyear{li2024learning} & 3D-Fauna  & I & \textit{U} & $\symbmesh$, \{$\symbscrew$+$\symbfd$\}, $\symbbb$, $\symbdqb$ & Quadruped Animals & PCK@0.1, KT-PCK@0.1\\

  
\midrule
\textcolor{blue}{} & \multicolumn{7}{c}{\textcolor{blue}{3D APE from Image \& Video - Multi-Camera-Views Methods}} \\
 \midrule
 18 & \citeyear{yao2019monet} & Monet  & I & \textit{SM} & $\symbposition$, $\symbshape$, $\symbvoxel$, $\symbsurfels$ & Dog, Monkey & PCK, Reprojection Err.\\

 19 & \citeyear{bala2020automated} & OpenMonkeyStudio  & I & \textit{S} & $\symbposition$, $\symbifm$, $\symbvoxel$, $\symbsurfels$ & Monkey & Median \& Mean Err. (cm), \\
  & & & & & & & Std (cm) \\

 20 & \citeyear{zhang2021animal} & GIMBAL  & V & \textit{S} & $\symbposition$, $\symblidar$, $\symbvoxel$, $\symbsurfels$ & Mouse & MPE (mm), RPA-MPE (mm)\\

 
 
 21 & \citeyear{dunn2021geometric} & DAN-NCE  & V & \textit{S} & $\symbposition$, $\symbifm$, $\symbvoxel$, $\symbsurfels$ &  Rats, Marmosets Mice, & PCK, Reconstruction Err.\\
 & & & & && Rat Pups \& Chickadees &\\
 
 22 & \citeyear{karashchuk2021anipose} & Anipose  & V & \textit{S} & $\symbposition$, $\symbifm$, $\symbvoxel$, $\symbsurfels$ &  Fly, Mouse & Joint Position Err. (mm)\\

  23 & \citeyear{muramatsu2022improving} & Muramatsu \etal  & V & \textit{S} & $\symbposition$, $\symbifm$, $\symbvoxel$, \symbdqb & AcinoSet (\citeyear{joska2021acinoset}) & RMSE, PCK, $l^2$ norm in pixels\\

  24 & \citeyear{ebrahimi2023three} & 3D-UPPER  & V & \textit{U} & $\symbposition$, $\symbsemanticsegmentation$, $\symbvoxel$, $\symbsurfels$ & Mice & RMSE, Scalar Product,  \\
       & & & & & & & Pearson’s Correlation \\

  
  25 & \citeyear{an2023three} & MAMMAL  & V & \textit{S} & \{$\symbposition$, $\symbmesh$\}, \symbifm, $\symbvoxel$, $\symbsurfels$ & Pigs, Mouse, Dogs & MPJPE, IoU,\\
 & & & & & & & Avg. Err. on each keypoint (cm) \\

  26 & \citeyear{han2024multianimal} & SBeA  & V & \textit{SF} & $\symbposition$, \{\symbifm+$\symbdb$\}, $\symbvoxel$, $\symbsurfels$ & Mice, Dogs, Birds & RMSE (pixels), Reprojection Err.\\
  
\hline
\hline
\end{tabular}
\label{3D_Sing_Multi_view_APE_methods}
\end{center}
\end{table*}

\subsubsection{\textbf{Single Camera View}} \label{singleview}

Single-camera-view 3D APE methods reconstruct 3D animal poses from monocular images and videos, which presents a challenge due to the inherent ambiguity in deriving depth from 2D projections. Existing methodologies within this domain can be classified into two categories: (1) skeleton-only methods, which target reconstructing the animal's skeletal structure. (2) Animal Mesh Recovery (AMR), which recovers the animal's entire mesh.

\textbf{Skeleton-only methods} are dedicated to estimating the 3D coordinates of animal joints for a skeletal representation as their result. Inspired by the advancements in 2D APE, the paradigm of inferring 3D poses from estimated 2D poses, termed \textbf{2D-to-3D lifting}, has garnered considerable attention. This approach typically leverages strong priors for predicting 3D joint locations from a single-camera view. A notable example is the LiftPose3D framework \citep{gosztolai2021liftpose3d}, which trains a supervised neural network utilizing 3D ground truth data to lift 2D poses from one camera. The training data is augmented with different camera angles and bone lengths, which allows the network to internalize camera parameters implicitly and accommodate variations in animal sizes. In contrast, recent innovations have endeavoured to estimate camera parameters and construct shape distributions explicitly \citep{hu20233d}. In this paper \citep{hu20233d}, 3D APE is formulated as an optimization problem utilizing a pre-defined mouse skeletal model. This 3D skeleton model imposes a strong prior, addressing the issue of occlusions and acting as a regularization on the over-parameterized joint space. These two methods hinge on extensive 3D annotated datasets, which are hard to apply to wild animals whose data are difficult to collect. 

Recent attempts have been made to address this challenge to eliminate the dependence on 3D ground truth through unsupervised 2D-to-3D lifting techniques. For instance, Dai \etal \citep{dai2023unsupervised} have introduced an unsupervised approach that, given a 2D pose, employs a pose generator network to generate a corresponding 3D posture, while a separate camera network estimates the camera rotation. During training, the generated 3D pose is subjected to random re-projections onto camera viewpoints to synthesize a new 2D pose. The synthesized 2D pose is then decomposed into a 3D pose and a camera rotation, based on which consistency losses are imposed in both 3D canonical poses and camera rotations for self-supervised training with geometric consistency.
Complementing this, Sosa \etal \citep{sosa2023horse} have proposed a similar methodology which relies on self-supervision with the same geometric consistency constraint for 3D pose estimation. Notably, while Dai \etal \citep{dai2023unsupervised} assume the availability of actual 2D poses for each input image, Sosa \etal \citep{sosa2023horse} demonstrate the feasibility of their approach with minimal assumptions, including unlabeled images and a small set of synthetic 2D poses. The minimalistic requirements and the availability of large volumes of unlabeled data suggest that this method \citep{sosa2023horse} could be readily generalized across diverse animal species. Another 2024 approach to automatically extract the 3D pose from single-camera-view images using only synthetic data for training is DigiDogs \citep{Digidogs_2024_WACV}. This method takes DINOv2 \citep{oquab2023dinov2} as the base architecture, which is a foundation model known for its robust pre-training capabilities and superior domain adaptation performance. These studies exploit synthetic data to predict 3D joint locations from 2D images, reinforcing the potential of synthetic augmentation in the domain of monocular 3D APE. 

Collectively, these methods in single-camera-view skeleton-only 3D APE underscore the dynamic evolution of the field, reflecting a shift towards more generalized and unsupervised learning paradigms that promise to enhance the versatility and applicability of pose estimation across a broader spectrum of animal species.

On the other hand, \textbf{AMR} for 3D APE from monocular images and videos is further categorized into model-based AMR and model-free AMR. The former predominantly utilizes parametric body models to infer the detailed 3D mesh of animals. The latter eliminates the need for a pre-existing parametric template model to recover 3D animal meshes.


One of the most well-known \textbf{model-based AMR} is the Skinned Multi-Animal Linear (SMAL) model \citep{zuffi20173d}, which provides a template for various animal classes based on 3D scanning of deformable toy animals. The SMAL model represents animals using a linear combination of shape bases, which are adapted to capture the diversity of animal forms, alongside a set of pose parameters to articulate the model. This framework serves as a foundation for many model-based AMR works. A notable enhancement of the SMAL model is SMALR \citep{Zuffi8578514}, which allows the recovery of detailed 3D shapes of unseen animals and creates textured 3D meshes. SMALST \citep{zuffi2019three} further amalgamates real textures with synthetically generated poses, shapes, and backgrounds. Beyond SMALST, the field has seen the development of breed-specific variants of SMAL, such as models for dogs \citep{ruegg2023bite, ruegg2022barc, biggs2020wldo, li2021coarse} and birds \citep{wang2021birds}. These models incorporate scaling parameters into the SMAL framework to enable breed-specific fitting. By using the pre-defined 3D shape and pose priors for the target animals, model-based AMR methods aid the animal shape reconstructions in ambiguous cases; the recovered mesh model further provides details for analyzing body size, posture, motion, behaviour, and health conditions. However, when we need to estimate the poses for new animal types without an existing template mesh, data collection for these unseen animals can be time-intensive, thus restricting model-based AMR's applicability to the diverse animal kingdom.


On the other hand, \textbf{model-free AMR} approaches \citep{kanazawa2018learning, goel2020shape, yang2022banmo, wu2023dove, yang2023reconstructing, wu2023magicpony, li2024learning, jakab2024farm3d} do not rely on pre-existing parametric body models, which offers greater flexibility when we have rare and diverse animals. Many of these methods attempt to learn 3D priors from large-scale internet images and videos in an unsupervised or self-supervised manner, focusing on one category at a time \citep{goel2020shape, wu2023dove, yang2023reconstructing, wu2023magicpony, jakab2024farm3d}. In contrast, 3D-Fauna (2024) \citep{li2024learning} introduces a pan-category deformable model capable of reconstructing any quadruped animal instance in a feed-forward manner, pushing towards a more generalizable and unconstrained understanding of animal forms from monocular images. However, the recovered 3D shapes of current model-free approaches often produce overly smooth surfaces, which do not preserve as much detail as the model-based methods; this severely limits their applicability for downstream applications.

\subsubsection{\textbf{Multiple Camera Views}} Multiple-camera-views 3D APE methods utilize multiple synchronized cameras to capture different perspectives of animals and reconstruct their 3D poses.

\begin{figure*}[!t]
\centering
\subfigure[An example of multiple camera views 3D APE method \citep{dunn2021geometric} that encompasses multi-views approach based on triangulation of 2D keypoint estimations.]{\includegraphics[width=0.39\linewidth, height=3.5cm]{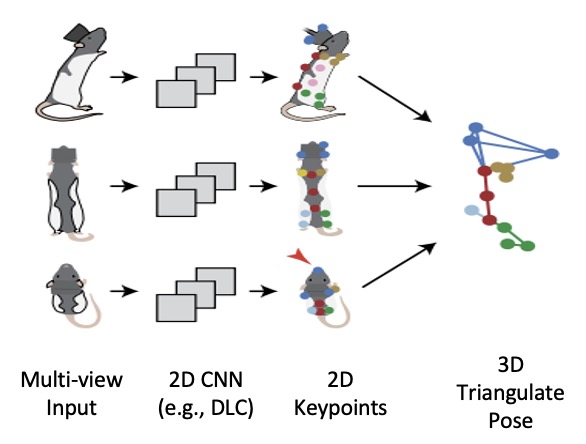}%
\label{fig:multiview}}
\hfil
\subfigure[An example of multiple camera views 3D APE method \citep{dunn2021geometric} that leverages multi-views geometric information during end-to-end training.]{\includegraphics[width=0.58\linewidth, height=3.6cm]{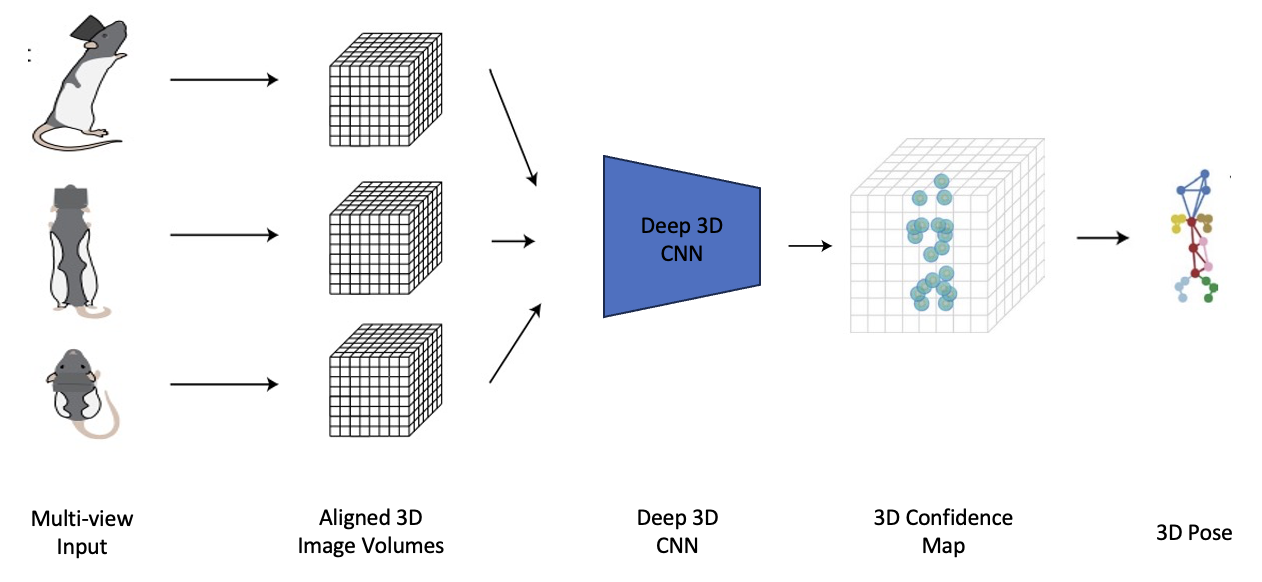}%
\label{fig:multiview2}}
\caption{Multiple camera views 3D animal pose estimation using ``DAN-NCE" \citep{dunn2021geometric}. \centering}
\label{fig:multiviewall}
\end{figure*}

The majority of the existing methods in multiple-camera-views 3D APE are designed to estimate the pose of a single animal in keypoints format \citep{yao2019monet, gunel2019deepfly3d, Nath2019, zhang2020multiview, bala2020automated, zhang2021animal, dunn2021geometric, karashchuk2021anipose, muramatsu2022improving, ebrahimi2023three}. In particular, early methods estimate the 2D poses from multiple independent camera views first; then, these 2D predictions are triangulated to 3D poses (\Cref{fig:multiview}). These post hoc triangulation approaches like \citep{Nath2019} work well when all the joints are visible in all views. However, they are not suited to handle occlusions, which is a scenario that always happens in animal naturalistic environments. To address these limitations, some methods apply skeletal constraints and filtering \citep{gunel2019deepfly3d, karashchuk2021anipose}. In detail, Anipose \citep{karashchuk2021anipose} benefits significantly from post-processing spatial-temporal filtering, especially in scenarios involving freely moving subjects where occlusion is frequent. However, these methods have yet to achieve robust 3D tracking of freely moving animals unless trained with large numbers of frames. A notable exception is OpenMonkeyStudio \citep{bala2020automated}, which can estimate the 3D macaques pose in extensive, unconstrained environments with a network of 62 high-resolution video cameras. However, these methods triangulate 3D locations from 2D estimates; their 2D nature limits their ability to capture and quantify behaviours in 3D accurately. To incorporate advances in 3D, some approaches integrate multiple-camera-views geometric data directly within end-to-end training pipelines (\Cref{fig:multiview2}). A notable system, DANNCE \citep{dunn2021geometric}, utilizes projective geometry to construct a 3D feature space from multiple-camera-views imagery and subsequently trains a 3D CNN to predict the positions of anatomical landmarks from the shared features across cameras. In addition, to tackle the data scarcity in 3D APE, 3D-Upper (2023) \citep{ebrahimi2023three} uses an unsupervised algorithm based on a Statistical Shape Model (SSM) to estimate 3D poses of freely moving animals; this SSM estimator is robust even when many poses (up to 50\%) are contaminated by outliers and missing data. The 3D-Upper APE method does not require labelled training data, which makes it highly scalable and more practical for applications in-the-wild or in situations where manual annotations are impractical. 

All of the aforementioned 3D multiple-camera-views approaches are limited to estimating the pose of a single animal, and they only provide poses in keypoints format, which does not preserve as much animal shape information as mesh-based APE methods. MAMMAL \citep{an2023three} and SBeA \citep{han2024multianimal} extend this to enable multi-animal 3D social pose estimation using multi-camera-views. In particular, MAMMAL is designed for large mammals and provides a full 3D surface mesh model compared to the sparse 2D keypoint methods. It captures fine-level surface geometries to understand animal motion involving interactions and occlusions in their complex social behaviours. Moreover, the cross-view matching algorithm further allows MAMMAL to reconstruct 3D poses robustly, even in the presence of keypoint occlusion. Triangulation-based methods are weak in such scenarios and may lead to incomplete or incorrect pose estimations. Another recent method, SBeA, further pushes the trend towards using a few-shot learning strategy to accurately estimate multi-animal 3D poses and identities with minimal manual annotation. SBeA introduces a novel bidirectional transfer learning strategy that enables zero-shot multi-animal identity recognition. It reuses models trained on single-animal data to identify multiple animals, thus minimizing extra identity annotations. The SBeA model employs multiview camera arrays and geometrical constraints to reconstruct 3D multiple animals' poses effectively, which is hard to attain through single-camera-view APE methods. Lastly, SBeA confers the capability for unsupervised behaviour classification by decomposing 3D trajectories into 2D representations and clustering social behaviour modules. This permits the discovery of subtle social behaviour phenotypes that supervised techniques might miss.

In summary, RGB-based APE methods have advanced significantly from 2011 to 2024. Early approaches, like DeepLabCut \citep{mathis2018deeplabcut}, focused on 2D keypoint-based supervised methods adapted from HPE for tracking animal joints in controlled environments. As research progressed, interest grew in 3D APE (particularly mesh-based methods), which capture more detailed pose and shape data for downstream tasks like tracking and behaviour analysis. However, the limited annotated data, and variations in animal species and environmental set-up in APE have initially restricted the progress of supervised APE methods. To address these, recent methods have shifted to weakly-supervised, self-supervised, semi-supervised, and unsupervised learning, using synthetic data, transfer learning, and domain adaptation to improve model generalization and tackle data scarcity. Additionally, given that many animals are social creatures, there is a trend toward multi-animal tracking from video data to understand their communication and social structures. These trends culminated in 2024 with the release of SuperAnimal \citep{superanimal_2024} (2D APE), MAMMAL \citep{an2023three} and SBeA \citep{han2024multianimal} (3D APE), which offer superior APE performance, generalization across species and diverse environments, reduced dependence on labelled data, and allow for tracking multiple animals simultaneously, setting a new benchmark in the field.


\section{Non-RGB-based Unimodal Animal Pose Estimation} \label{othersensors}

In the previous section, we comprehensively reviewed APE methods primarily based on RGB cameras. However, beyond RGB-based approaches, researchers have explored the use of other unimodal methods for analyzing animal behaviour, \Cref{Multi_Sensor_APE} summarises the latest APE methods using both non-RGB unimodal \citep{yigit2022wearable} and multi-modal approaches, which is discussed in detailed in \Cref{mulit_sensor_APE}. 

\textbf{IMU:}
IMU is a small, lightweight, low-cost device that measures an object's acceleration and angular velocity. Attaching multiple IMUs to limbs and the trunk makes it possible to capture the motion and orientation of the animal in 3D space. Recent work \citep{yigit2022wearable} presents a lameness detection and pose estimation method for horse walk and trot locomotion based on the acceleration and gyroscope measurements from IMUs. Initially, they train the network offline using IMU measurements as input sequences, with labelled data provided by the MoCap system. Once the long short-term memory (LSTM) model for gait event and lameness detection is trained, it embeds the IMU measurements into high-dimensional joint angles, which are then fed into the Gaussian Process Dynamic Model (GPDM) to learn a low-dimensional latent representation of the horse's limb motion. This learned latent manifold dynamics is parameterized with the gait phase variable to predict the horse's 24 limb joints in real time. 

Currently, non-RGB unimodal APE methods have yet to be widely adopted. However, using multi-modality approaches, such as combining RGB cameras with other sensors, has gained more popularity, as discussed in \Cref{mulit_sensor_APE}.



\begin{table*}[ht]
\renewcommand{\arraystretch}{1}

\caption{Non-RGB-based unimodal APE methods and multi-modal APE methods. All the symbols and column meanings are defined in \Cref{2D_Image_APE_methods}, \Cref{2D_APE_Video_methods} and \Cref{3D_Sing_Multi_view_APE_methods}, and any additional symbols are defined here: $\symbef$ is GPDM architecture, $\symbautodecoder$ is Hierarchical Gaussian Process Latent Variable Model (H-GPLVM), \symbrr is Extended Kalman Smoother (EKS), $\symbfdm$ is Context Models, $\symboff$ is CLIP. $\symbsdf$ is ego-view. Ther. is an abbreviation for thermal, and Lang. is an abbreviation for language. The method citation is available in column Year.}
\begin{center}
\small
\setlength{\tabcolsep}{1pt}
\begin{tabular}{c|c|l|c|c|c|c} 
\hline
\textbf{Sl} & \textbf{Year} & \textbf{Methods} & \textbf{Sensor} & \textbf{Features} & \textbf{Evaluated On}& \textbf{Metric}\\
    
    \midrule
    \textcolor{blue}{} & \multicolumn{6}{c}{\textcolor{blue}{Non-RGB-based Unimodal APE Method: IMU-based}} \\
    \midrule 
    1 & \citeyear{yigit2022wearable} & Yigit \etal  & IMU & $\symbposition$, $\symbef$, $\symbbb $, \{$\symbsurfels$, $\symbdqb$\} & Horse & Mean, Std, VR, IAE \& IEV \\ 

    \midrule
    \textcolor{blue}{} & \multicolumn{6}{c}{\textcolor{blue}{Multi-Modal APE Methods}} \\
    \midrule 

    2 & \citeyear{patel2017tracking} & Patel \etal  & RGB+GPS+IMU & $\symbposition$, \symbrr, \symbsdf, \symbdqb & Cheetah & Qualitative for 3D \\
    &&&  & & & (video observations) \\

    \hline
    3 & \citeyear{RGBD_Dogs_2020_CVPR} & Kearney \etal  & RGBD & $\symbmesh$, \{$\symbifm$+$\symbautodecoder$\}, $\symbbb$, $\symbsurfels$ &  BADJA (\citeyear{BADJA_biggs2018creatures}), & PCK, MPJPE, PA-PCK, \\
    & & & & & RGBD-Dog (\citeyear{RGBD_Dogs_2020_CVPR}) & PA-MPJPE, PCK 3D,\\
    &&&  & & &   PA-PCK 3D \\

    4 & \citeyear{RGBD_SSM_2023} & Luo \etal  & RGBD & \{$\symbposition$, $\symbmesh$\}, $\symbei$, $\symbvoxel$, $\symbsurfels$ & Cattles, Pigs (\citeyear{RGBD_dataset_2020_Ruchay_et_al}) & MAE, MRE (\citeyear{RGBD_SSM_2023}) \\


    \hline
    5 & \citeyear{patel2023animal} & OptiPose  & RGBD+Ther. & $\symbposition$, $\symbfdm$, $\symbvoxel$, $\symbsurfels$ & Rat7M (\citeyear{dunn2021geometric}),  & PCK, MPJPE, \\
    &&& Infrared & & AcinoSet (\citeyear{joska2021acinoset}), & Reprojection Error  \\
    &&& Infrared & & Rodent3D (\citeyear{patel2023animal}) &  \\

    \hline
    6 & \citeyear{WildPose2024} & WildPose & RGB+LiDAR & $\symbposition$, N/A, $\symbbb$, $\symbdqb$ & Eagle, Giraffe, Lion  & Mean \& Std \\
    \hline
    
    7 & \citeyear{li2022sound} & Sound of Motion  & RGB+Acoustic & $\symbmesh$, $\symbei$, $\symbbb$, \symbdqb & Horse & PCK, IoU, PA-MPJPE \\
    
    8 & \citeyear{CLHOP_Audio_video_2024} & CLHOP  & RGB+Acoustic & $\symbmesh$, \{$\symbei$+$\symbifm$\}, $\symbbb$, \{$\symbsurfels$,$\symbdqb$\} & Horse (\citeyear{CLHOP_Audio_video_2024}) & PCK@0.1, IoU, PA-MPJPE \\

    \hline
    9 & \citeyear{language_zhang2023clamp} & CLAMP  & RGB+Lang. & $\symbposition$, $\symboff$, $\symbbb$, $\symbdqb$ &  Animal Pose (\citeyear{cao2019cross}), & AP, AR\\
    &&&  & & AP-10K (\citeyear{yu2021ap}) &  \\

    10 & \citeyear{AmadeusGPT_2023} & AmadeusGPT  & RGB+Lang. & $\symbposition$, $\symbifm$, N/A, $\symbsurfels$ & Mouse \& others (\citeyear{AmadeusGPT_2023}) & F1 Score \\
    
    11 & \citeyear{AWOL_RGB_Lang_2024} & AWOL  & RGB+Lang. & $\symbmesh$, $\symbei$, N/A, $\symbdqb$ & StanfordExtra (\citeyear{biggs2020wldo}),   & CLIP score, Qualitative, \\ 
    &&&  & & 3D Trees & Perceptual Study \\

\hline
\hline
\end{tabular}
\label{Multi_Sensor_APE}
\end{center}
\end{table*}

\section{Multi-Modal Animal Pose Estimation}

Researchers have begun to explore the integration of multi-sensor and multi-modality to improve the accuracy, robustness, and ecological validity of APE algorithms \citep{patel2017tracking, RGBD_Dogs_2020_CVPR, li2022sound, RGBD_SSM_2023, patel2023animal, language_zhang2023clamp, AmadeusGPT_2023, WildPose2024, 10207590}. In this section, we examine the motivation behind multi-modal systems, review existing multi-modal HPE and APE techniques, discuss how they tackle the limitations in unimodal APE methods, and highlight applications demonstrating the potential of integrating diverse sensors and modalities for APE, as shown in \Cref{Multi_Sensor_APE}.

\subsection{Multi-Modal System in Human Pose Estimation} \label{multisensorhpesection}

To inspire the use of multi-modal systems for APE, we draw insights from recent multi-modal HPE studies. For instance, \citep{zheng2022multi, zanfir2023hum3dil} combine data from visual RGB cameras with depth information obtained from RGBD and LiDAR sensors in autonomous driving to increase the robustness of 3D HPE in complex environments. The mRI dataset \citep{an2022mri} integrates mmWave radar, RGBD cameras, and IMUs and provides a benchmark to support 3D HPE sensor fusion research. Another study \citep{an2022mri} introduces the mRI dataset, a multi-modal dataset for 3D HPE that integrates data from mmWave radar, RGBD cameras, and IMUs. These studies highlight the benefits of multi-modal integration in HPE and further inspire us to explore similar approaches for APE.

\subsection{Motivation for Multi-Modal System in Animal Pose Estimation}

Unimodal APE faces challenges like occlusions and limited perspectives, which multi-modal systems can overcome by combining complementary data to improve accuracy and understanding of animal behaviour \citep{10207590}. For example, multi-modal systems enhance APE resilience by compensating for sensor failures, such as using thermal in low-light conditions when RGB cameras struggle. In addition, RGB-based cameras excel in capturing detailed visual information but may falter with occlusions, while some mmWave sensors can see through walls, and LiDAR can provide valuable 3D spatial data; these significantly help us to expand the coverage area and reduce blind spots. Beyond visual data, integrating IMUs and audio sensors further enriches animal behaviour analysis by capturing motion and auditory cues, which makes multi-modal systems essential for a more robust APE and a better understanding of dynamic environments.

\subsection{Existing Multi-Modal Animal Pose Estimation Techniques}\label{mulit_sensor_APE}

In recent years, there has been growing interest in the development of multi-modal APE systems \citep{RGBD_Dogs_2020_CVPR, RGBD_SSM_2023, patel2017tracking, li2022sound, patel2023animal, language_zhang2023clamp, AmadeusGPT_2023, WildPose2024, CLHOP_Audio_video_2024, AWOL_RGB_Lang_2024}. It involves combining information from different sensors to derive a unified representation of animal pose. Compared to the significant progress in unimodal APE, there is limited research on using multiple sensors and modalities for enhanced accuracy in APE, and we aim to review all of the existing multi-modal APE studies in this section.

\textbf{Animal-borne cameras + GPS + IMUs (Ego-view):}
One notable example of a multi-modal system using animal-borne cameras, GPS, and IMU is the work on tracking the cheetah tail \citep{patel2017tracking} in ego-view. Ego-view 3D APE, though less explored, presents a novel perspective by placing cameras on the animals themselves, offering a first-person viewpoint. In this study \citep{patel2017tracking}, researchers attached camera and sensor equipment to cheetahs to obtain motion data from the animal's viewpoint. Using kinematic models, they can estimate the 3D pose of the cheetah's tail, including its position, orientation, and movement in 3D space. This research demonstrates the potential of multi-modal ego-view APE methods in capturing the animal's pose and movement directly from the animal's perspective, which helps us gain insights into the animal's behaviour, locomotion, and agility so we can have a better understanding of the cheetah's hunting strategies, balance control, and communication signals. Please note that the ego-view 3D APE study is currently limited, and the cheetah tail tracking study opens a door for it.

\textbf{RGB + Depth:}
One promising sensor modality is the RGBD-based pose estimation, as exemplified by \citep{RGBD_Dogs_2020_CVPR}. The pose estimation model is a two-stage approach for predicting 3D canine poses from RGBD data. In the first stage, a stacked hourglass network is employed to predict 2D joint locations from RGB images, with multiple stacked layers that allow for iterative refinement of joint positions. To generalise the model across various canine shapes and sizes, this study trains the network on both real and synthetic data. In the second stage, a Hierarchical Gaussian Process Latent Variable Model (H-GPLVM) is used to lift the predicted 2D joint positions to 3D, while ensuring the output poses are anatomically plausible by imposing constraints on the possible configurations. The H-GPLVM leverages the structure of the canine skeleton to refine the pose estimations, which further reduces errors and improves robustness in complex poses or challenging viewpoints. This model can extract dog poses effectively even when there is significant variability in appearance and motion. In 2023, there is another RGBD-based APE approach \citep{RGBD_SSM_2023} that enhances livestock body measurements. It first preprocesses point cloud data to detect 3D keypoints, then aligns a template mesh (\ie a statistical shape model) with the scan data, and uses non-rigid registration and poses normalisation to standardise the animals' poses into a canonical form. These help to address the common challenges such as noise, missing data, and inconsistencies caused by livestock movement. Even though RGBD is a promising approach, its popularity is limited by dataset availability and hardware constraints; hence, it has not yet been widely adopted within the APE community compared to RGB-based methods.

\textbf{RGB + Depth + Thermal infrared:} The OptiPose study \citep{patel2023animal} introduces a novel approach for animal pose tracking that leverages a multi-modal dataset and token-based pose optimisation. To capture accurate APE, researchers collected a 3D multi-modal dataset comprising RGB, depth, and thermal infrared data. RGB data provides high-resolution colour imagery that helps identify detailed animal features, allowing for better 2D keypoint extraction. Depth data provides spatial information about the distance between the camera and the animal, which is important for reconstructing the 3D pose from multiple 2D views. Finally, thermal infrared data measures the heat emitted by the animal, which is particularly useful in low-light or occluded conditions where RGB data struggled. They propose a token-based optimisation method that uses multi-modal data to refine the pose estimation results. The study demonstrates the potential of integrating multiple sensor modalities for improved animal pose tracking and motivates further exploration of multi-modal approaches in animal behaviour analysis.

\textbf{RGB + LiDAR:}
A novel approach to APE uses LiDAR and RGB sensors, as exemplified by the WildPose \citep{WildPose2024} system. Unlike traditional methods that rely on RGB images, WildPose utilises a combination of a zoom-lens camera and a solid-state LiDAR to capture both 2D and 3D data of animals in their natural habitats. The WildPose system is non-invasive and can capture creatures in a long-range distance (ranging from 2-200 meters) without disturbing the animals. Moreover, LiDAR adds depth information about the animals and their surrounding environments to provide a more comprehensive understanding of the animal’s morphology and movements. Despite these advantages, the high cost and complexity of LiDAR systems may limit their accessibility and scalability. Additionally, the accuracy and quality of the LiDAR data can be affected by environmental factors such as weather conditions and lighting.

\textbf{RGB + Acoustic:} The Sound of Motion \citep{li2022sound} and CLHOP \citep{CLHOP_Audio_video_2024} present a multi-modal approach for estimating horse motion using video and audio data. Audio features are extracted from sounds like horse hoof strikes and are processed in parallel with visual features. These audio features are converted into Mel spectrograms and passed through a network that helps predict 3D pose parameters. These studies highlight the value of incorporating audio to capture fine-grained details of animal movements. Apart from APE, audiovisual data have also been applied in other animal motion analyses such as horse behaviour detection \citep{nunes2021horse} and primate action recognition \citep{maxbainwildprimates}.

\textbf{RGB + Language:}
As we advance, the APE trend is moving towards using vision-language-based APE methods \citep{language_zhang2023clamp, AWOL_RGB_Lang_2024}. The CLAMP (2023) \citep{language_zhang2023clamp} method uses language to enhance learning processes through the description provided, thus increasing the capability of generalising models about new and unseen animal poses. Following that, the AWOL (2024) \citep{AWOL_RGB_Lang_2024} approach creates a model that takes input from Contrastive Language-Image Pretraining (CLIP) encodings of either text or images and maps them to 3D shape parameters. This learns the mapping between the CLIP latent space and the shape space of SMAL, allowing it to generate realistic 3D animal mesh as a final output. This exploits the strengths of vision-language models in blending visual and textual information into a more robust framework, advancing the APE field of work.

In conclusion, the above studies provide insights and inspirations for future work, which shows a trend towards integrating multiple sensors and modalities, such as RGB, depth, IMUs, LiDAR, thermal, acoustic, and language, to enhance the accuracy, robustness, and contextual understanding of APE.

\section{Future Work}

\subsection{\textbf{Can we reduce or eliminate the effort required for ground-truth annotation in APE using unsupervised learning and meta-learning?}}

Unsupervised learning and meta-learning hold great potential to significantly reduce or even eliminate the effort required for ground-truth annotation in APE. It will be worth exploring how \textbf{unsupervised learning} models can leverage inherent structures in the data, such as consistency across multiple views, temporal coherence in video sequences, or self-supervised signals like motion cues. By exploiting such patterns, unsupervised APE models can hopefully learn from vast amounts of unlabeled data, which is particularly useful in wildlife observation where labelled data is scarce. Future APE work will also benefit from \textbf{meta-learning} capabilities. Meta-learning focuses on teaching models to learn how to learn, enabling them to generalise quickly to new species or environments with minimal labelled data. In APE, meta-learning will be able to train models on a limited set of labelled animal poses to generalise to entirely new species or different environmental conditions using only a small number of annotated examples. Lastly, future work will also combine both approaches to leverage large amounts of unlabeled data for initial training and then rapidly adapt to specific tasks with minimal supervision. This hybrid approach would dramatically reduce the need for extensive annotations while improving the model's robustness and generalisation to new tasks.

\subsection{\textbf{How can we extend multi-modal frameworks to perform multiple-animal pose estimation in the scene?}}

Currently, the use of multi-modal frameworks for a single animal is limited, and for multi-animal, minimal work exists (\ie OptiPose \citep{patel2023animal} that can perform APE on two rodents from a multi-modal system). However, the future work will include \textbf{multi-modal multi-animals} pose estimation in-the-wild for various species. Going forward, we will see a fusion of sensors and modalities for handling occlusions and provide additional context when animals interact closely. For example, depth data will help to differentiate between overlapping animals, while thermal data will assist in distinguishing animals from their surroundings in low-light conditions. In addition, we may also use multi-modal data to combine animal appearance and other information such as their shape, weight, and gaits together to resolve \textbf{individual animal identity}, which is very important for agriculture and conservation. In future, work will move towards \textbf{scene object interaction and reconstruction}. By using multi-modal data, we will reconstruct both the animals and their surrounding objects by understanding context-specific interactions, such as animals navigating around obstacles, interacting with other animals, or using objects (\eg chimpanzees nut cracking in-the-wild). For instance, LiDAR or depth sensors will capture the 3D geometry of the environment, and when combined with RGB data, the system will reconstruct both the animal's poses and the objects they interact with. The future will also move towards having \textbf{real-time processing}, which will allow us to track pose and behaviour in real-time, enabling effective health monitoring to protect endangered species.

\subsection{\textbf{Can we perform language-driven animal articulation with advanced APE methods?}}

With the rise of models like CLIP, the trend is moving towards more vision-language hybrid approaches within the CV community. So far, there is no work available where it is possible to control and articulate 3D animal models using natural language commands in an unsupervised manner, but this will likely happen in the future. By integrating language-driven models with advanced APE techniques, we can achieve sophisticated \textbf{text-based control for animal poses}. Similar to how text prompts can control image generation in models like CLIP, future APE methods could allow users to articulate an animal's pose using natural language instructions. For instance, commands like ``raise the horse's front right leg" or ``make the chimpanzee sit" could be translated into precise joint movements or skeletal adjustments in a 3D model. Moreover, mesh-based APE methods, when integrated with language models, could allow for real-time adjustments in the animal’s anatomy and surface details with commands like ``increase muscle tension" or ``broaden the horse's stance" and this could instantly alter both the pose and physical appearance of the 3D mesh model in real-time. These advancements will be beneficial in applications like virtual reality or animations, where controlling animal movements dynamically with text could simplify workflows. It also has the potential for behavioural simulations; for example, researchers or animators can generate lifelike animal behaviours through high-level language inputs without manually defining joint movements.

\section*{Declarations}
We thank Professor Andrew Zisserman for his excellent suggestions and for reviewing the paper. We would also like to thank Jiaxing Zhong, Ta-Ying Cheng, Xinyu Hou, Qian Xie, Shaokai Ye, Daniel Schofield, Eldar Insafutdinov, Niki Amini-Naieni, Tomas Jakab and Anjun Hu for insightful technical discussion. We would like to thank \citep{3D_Recon_Survey_2024_Chris, deb2023} for using some of their symbols, templates and scripts. Additionally, Qianyi would like to thank Jian Zhou for his loving support. During this project, Qianyi Deng was supported by the UKRI's EPSRC ACE-OPS grant (EP/S030832/1), and Oishi Deb was supported by the UKRI's EPSRC AIMS CDT grant (EP/S024050/1).

\bibliography{sn-bibliography}

\end{document}